\documentclass[journal]{IEEEtran}

\ifCLASSINFOpdf
\else
\fi
\usepackage{cite}

\usepackage{multirow}
\usepackage{amsmath}
\usepackage{amssymb}
\newcommand{\norm}[1]{\left\lVert#1\right\rVert}

\usepackage[pdftex]{graphicx}
\graphicspath{{../pdf/}{../jpeg/}}
\DeclareGraphicsExtensions{.pdf,.jpeg,.png}

\usepackage{caption2}
\usepackage{stfloats}
\usepackage{bm}

\usepackage{color}

\usepackage{algorithm} 
\usepackage{algpseudocode}

\begin{document}
%
\title{A Lightweight Recurrent Learning Network for Sustainable Compressed Sensing}
%
%
%

\author{Yu Zhou, \IEEEmembership{Member, IEEE},
        Yu Chen,
        Xiao Zhang, \IEEEmembership{Member, IEEE}, Pan Lai, Lei Huang, \IEEEmembership{Senior Member, IEEE} and
        Jianmin Jiang

\thanks{This work was supported in part by the Natural Science Foundation China (NSFC) for Distinguished Young Scholars under Grant 61925108, and in part by the Joint fund of the National Natural Science Foundation of China and Robot Fundamental Research Center of Shenzhen Government under Grant U1913203, in part by the Natural Science Foundation China (NSFC) under Grant 61702336, 61902437, 62032015 and 61620106008, and in part by the Shenzhen Fundamental Research Program under Grant JCYJ2020010911041013.
\emph{(Corresponding author: Xiao Zhang.)}}
\thanks{Yu Zhou, Yu Chen and Jianmin Jiang are with the College of Computer Science and Software Engineering, Shenzhen University, Shenzhen 518060, China (email: yu.zhou@szu.edu.cn, 2070276225@email.szu.edu.cn, jianmin.jiang@szu.edu.cn).}
\thanks{Xiao Zhang and Pan lai are with the Department of Computer Science, South-Central Minzu University, Wuhan 430074, China (e-mail: xiao.zhang@my.cityu.edu.hk, plai1@ntu.edu.sg).}
\thanks{Lei Huang is with the College of Electronics and Information Engineering, Shenzhen University, Shenzhen 518060, China (e-mail: lhuang@szu.edu.cn).}}

%
%

\markboth{Journal of \LaTeX\ Class Files,~Vol.~14, No.~8, August~2015}%
{Shell \MakeLowercase{\textit{et al.}}: Bare Demo of IEEEtran.cls for IEEE Journals}
%



\maketitle

\begin{abstract}
Recently, deep learning-based compressed sensing (CS) has achieved great success in reducing the sampling and computational cost of sensing systems and improving the reconstruction quality. These approaches, however, largely overlook the issue of the computational cost; they rely on complex structures and task-specific operator designs, resulting in extensive storage and high energy consumption in CS imaging systems. In this paper, we propose a lightweight but effective deep neural network based on recurrent learning to achieve a sustainable CS system; it requires a smaller number of parameters but obtains high-quality reconstructions. Specifically, our proposed network consists of an initial reconstruction sub-network and a residual reconstruction sub-network. While the initial reconstruction sub-network has a hierarchical structure to progressively recover the image, reducing the number of parameters, the residual reconstruction sub-network facilitates recurrent residual feature extraction via recurrent learning to perform both feature fusion and deep reconstructions across different scales. In addition, we also demonstrate that, after the initial reconstruction, feature maps with reduced sizes are sufficient to recover the residual information, and thus we achieved a significant reduction in the amount of memory required. Extensive experiments illustrate that our proposed model can achieve a better reconstruction quality than existing state-of-the-art CS algorithms, and it also has a smaller number of network parameters than these algorithms. Our source codes are available at: https://github.com/C66YU/CSRN.

\end{abstract}


%
\IEEEpeerreviewmaketitle

\section{Introduction}
%
%
%
%
\IEEEPARstart{C}{ompressed} sensing (CS) theory \cite{1614066} is an emerging paradigm for signal sampling and reconstruction that demonstrates that sparse signals can be recovered from linear random measurements with a high probability at sample ratios lower than that required by Nyquist sampling theory. In traditional signal processing methods, to make data storage and transmission more convenient, the sampled signal will be further compressed, with part of the data discarded, resulting in the waste of sampling resources. On the contrary, compressed sensing has the advantage of being able to compress and sample signals simultaneously, which greatly reduces the sampling and computational cost of the sensing system. Therefore, compressed sensing has attracted great attention and been widely used in many applications, such as imaging systems \cite{7778203}, medical scanners \cite{7486011}, etc.

Due to the high efficiency and the many potential applications of CS theory, many algorithms that exploit the potential of compressed sensing for images, such as those reported in \cite{4288604,5652744,5453522,6814320,7406719,7076640,6232453}, have been explored in the literature. Although these algorithms can recover the original signal well, their common weakness is that all these algorithms attempt to solve convex optimization problems in an iterative manner even though the iterative process is often computationally intensive and thus slows down the reconstruction speed, which limits the real-time performance and further application of these algorithms. 

In recent years, deep learning has achieved great success in a range of computer vision tasks due to its advanced performance and high-speed inference abilities; these tasks include image classification \cite{10.5555/2999134.2999257}, super resolution \cite{7115171} and image restoration \cite{9134805}. Hence, some CS algorithms based on deep learning have also been proposed\cite{9025255,8233175,ma2022deep}. Compared with traditional CS algorithms, deep learning-based CS algorithms reconstruct images faster thanks to their non-iterative nature. Deep learning-based algorithms take a significant amount of time to learn appropriate model parameters during the training process, but when inferences are being made, the original signal can easily be recovered through computationally efficient matrix multiplications and nonlinear operations. Additionally, the strong learning ability of deep neural networks also ensures that such algorithms can capture the characteristics of signals and provide a better reconstruction quality. 

Despite the positive progress that has been achieved, existing deep learning-based CS algorithms still have limitations. On the one hand, current approaches do not fully explore and utilize the the potential power of feature extraction and learning in neural networks, which results in limited reconstruction quality improvement. On the other hand, to achieve high-quality reconstructions, most current methods employ very complex network structures with problem-specific operators, leading to a huge number of parameters to be optimized and extensive memory and energy costs, which impede the use of such algorithms on devices with limited computational resources; these methods cannot meet the demand for sustainable artificial intelligence in our modern society.
It is worth mentioning that although parameter pruning techniques such as neural architecture search (NAS) can help to reduce the number of parameters \cite{van2019evolutionary,yang2020ista}, the reconstruction quality is often limited due to the fact that, on the one hand, the manually designed complex network architecture fails to provide sufficient space for quality improvement, and on the other hand, unlike classification tasks, CS aims to reconstruct the data as accurately as possible; reducing the number of network parameters will ultimately degrade the reconstruction quality.

In recent years, recurrent learning has been introduced into vision tasks due to its excellent image-modeling capabilities. For example, DuDoRNet \cite{zhou2020dudornet} shows that recurrent learning can better avoid overfitting by directly optimizing restoration networks in dual domains, and it can reconstruct high-quality images with a low number of parameters. PolSAR \cite{ni2020random} is a novel recurrent learning strategy that learns polarization and spatial features from adjacent pixel sequences, which further reduces intraclass and interclass misclassifications. In addition, Adler et al. propose a sustainable CS approach that is mostly suitable for processing very high-dimensional images and videos: it operates on local patches, employs a low-complexity reconstruction operator and requires significantly less memory to store the sensing matrix \cite{adler2017block}. Reddy et al. present an effective lightweight data-reduction method by investigating the performance of CS-based and partial discrete cosine transform (DCT)-based compression methods \cite{reddy2021lightweight}. However, these methods still have some drawbacks; for example, they are not flexible enough, and the reconstruction quality depends on the complex design of the sampling matrix.

In smart cities, data come from different sources, and data collected and integrated through IoT devices are important. In this context, finding a balance between the accuracy of the data collection and the efficient use of limited resources through lightweight and efficient methods is critically important. In this paper, to realize a sustainable CS model, we focus on designing a lightweight but effective recurrent learning network called the compressed sensing recurrent network (CSRN). The proposed CSRN is based on a classical network structure that contains three sub-networks, namely, the sampling sub-network, initial reconstruction sub-network and residual reconstruction sub-network. While the sampling sub-network uses convolutional layers to perform the CS sampling process for original images, the initial reconstruction sub-network is used to recover an initial reconstructed image via linearly mapping measurements; the residual reconstruction sub-network is responsible for using nonlinear mapping to obtain high-frequency information and hence enhance the reconstructed details to provide higher-quality refinements. The final reconstructed image is obtained by integrating the initial reconstructed image and the residual image. Compared with existing approaches, our proposed CSRN differs in the following ways:

(1) We propose a flexible initial reconstruction sub-network consisting of basic recovery blocks to obtain initial images at different sample ratios, which can greatly reduce the number of network parameters.

(2) We propose a novel recurrent residual fusion module (RRFM) under the framework of recurrent learning to take full advantage of features from different scales.

(3) We employ a feature compression strategy in the residual image reconstruction process to further improve the efficiency of the reconstruction process.

The remainder of this paper is organized as follows. First, CS and the related algorithms are described in Sec. \uppercase\expandafter{\romannumeral2}. Next, we describe the proposed method in Sec. \uppercase\expandafter{\romannumeral3}. In Sec. \uppercase\expandafter{\romannumeral4}, we describe the extensive experiments that we performed and illustrate that our proposed algorithm outperforms a number of representative state-of-the-art methods. Finally, Sec. \uppercase\expandafter{\romannumeral5} provides concluding remarks.

\section{Related Work}
\subsection{Compressed Sensing}
Compressed sensing is an emerging paradigm for signal processing that can be used to reconstruct sparse signals from a smaller number of linear measurements and thus greatly improve the efficiency and effectiveness of a sensing system. Mathematically, the sampling process of CS can be formulated as follows:
\begin{equation}
    \textbf{y} = \bm{\Phi} \textbf{x},
\end{equation}
where $\textbf{x} \in R^N$ denotes the original signal, $\textbf{y} \in R^M$ is the linear measurement and $\bm{\Phi} \in R^{M\times N} (M \ll N)$ is the sampling matrix. In most cases, $\textbf{x}$ can be recovered by solving the following equation:
\begin{equation}
    \min_{\textbf{x}} \norm{\bm{\Psi} \textbf{x}}_0, s.t.\; \textbf{y}=\bm{\Phi} \textbf{x},
\end{equation}
where $\bm{\Psi}$ is the sparse transform matrix, and $\norm{\cdot}_0$ denotes the $l_0$ norm. However, solving $l_0$ optimization is an NP-hard problem; the solving process has a high complexity and may be unstable when noise interferes with the sampling process. The $l_1$ norm can be used to replace the $l_0$ norm to alleviate the above problems. Therefore, the signal can be recovered by solving 
\begin{equation}
    \min_{\textbf{x}} \norm{\bm{\Psi} \textbf{x}}_1, s.t.\; \textbf{y}=\bm{\Phi} \textbf{x}.
\end{equation}
Eq. (3) is a convex problem, and its solution is equivalent to
\begin{equation}
    \min_{\textbf{x}} \norm{\textbf{y} - \bm{\Phi} \textbf{x}}_2^2 + \lambda \norm{\bm{\Psi} \textbf{x}}_1,
\end{equation}
where the first part and the second part are the fidelity term and regularization term, respectively, and $\lambda$ is a parameter that is used to control the regularization strength.

\begin{figure*}[h]
    \centering
    \includegraphics[width=0.95\textwidth]{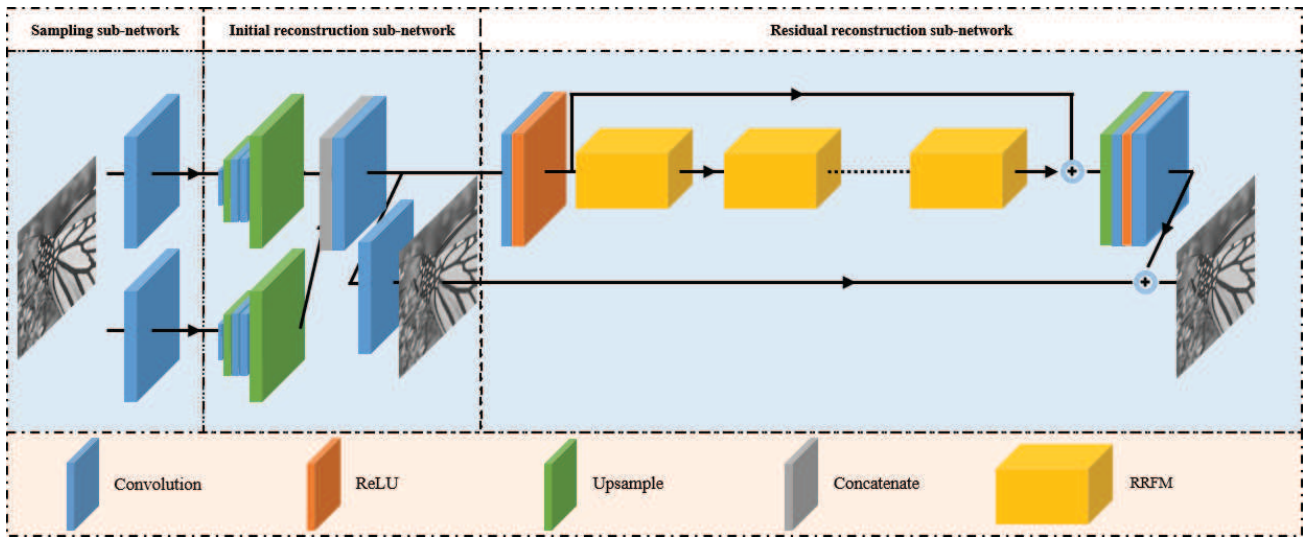}
    \caption{Network architecture of the proposed CSRN when the sample ratio is 0.2. For the sampling process, two $Conv(102, 32, 32)$ layers are applied to mimic image sampling. For reconstruction, the initial reconstruction sub-network with a two-layer structure is used to reconstruct the initial image, and then the residual reconstruction sub-network is used to recover the residual image, which will be added to the initial image to obtain the final reconstructed image.}
    \label{network_structure}
\end{figure*}

\subsection{Plain Deep Learning Network for Image Compressed Sensing}

Inspired by the success of deep learning in computer vision tasks, many deep learning-based algorithms have been developed to overcome the shortcomings of traditional optimization-based algorithms. 

Deep learning-based methods can be divided into two categories according to the network design. The first type of network is the plain network. This type of method attempts to directly learn a reconstruction algorithm from huge amounts of data. In \cite{7447163}, Mousavi et al. introduce deep learning into image CS and use a stacked denoising autoencoder to reconstruct images from measurements. Inspired by SRCNN, Kulkarni et al. propose ReconNet, which is based on a Convolutional neural network (CNN), in \cite{7780424}; a fully connected layer and multiple convolutional layers are used to reconstruct images from measurements. In \cite{YAO2019483}, Yao et al. propose Dr\textsuperscript{2}-Net, which uses a linear mapping to reconstruct a preliminary image and then uses residual learning to further improve the image quality. However, Dr\textsuperscript{2}-Net only learns the intra-block information and suffers from blocking artifacts. To solve this problem, in \cite{9199540}, DPA-Net is introduced; it has a dual path scheme to recover the image's structure and texture, respectively, and an attention mechanism is also integrated into the network to boost the texture recovery. Moreover, Shi et al.
propose CSNet, which uses residual networks to learn the
map between the measurements and reconstructed image,
to obtain fast end-to-end image reconstruction \cite{8019428}. DPA-Net and CSNet both alleviate blocking artifacts to a certain extent. 
However, these types of completely data-driven models have poor interpretability due to their black-box characteristics.

\subsection{Deep Unfolding Network for Image Compressed Sensing}
The second category is the deep unfolding network; this type of network unfolds the iterative process of optimization-based methods using a neural network, which makes the network more interpretable. In \cite{8578294}, Zhang et al. propose ISTA-Net, which casts ISTA into a neural network form to combine the advantages of optimization-based methods and deep learning-based methods. AMP-Net \cite{9298950} is proposed to unfold the iterative process of an approximate message mapping algorithm to reconstruct images. Further, AMP-Net uses a learnable deblocker to remove block artifacts to enhance the reconstruction quality. In \cite{9467810}, COAST is proposed to process an arbitrary sampling matrix using one deep unfolding model through random projection augmentation and a controllable proximal mapping module. Although these deep unfolding networks achieved good interpretability, it is still difficult to obtain a good reconstruction quality at a low sample ratio due to the network design.

\begin{figure}
    \centering
    \includegraphics[width=3.0in]{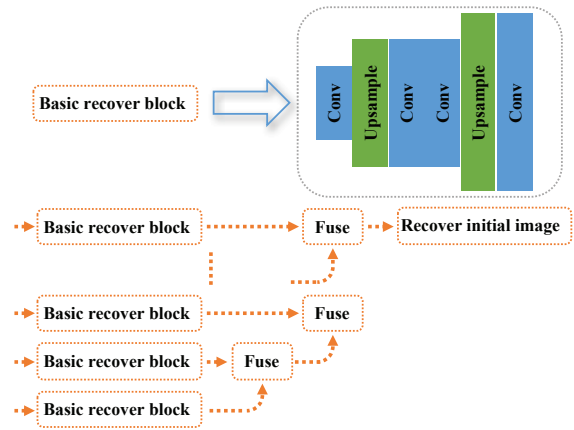}
    \caption{The structure of the initial reconstruction sub-network when the sample ratio is greater than 0.1.}
    \label{initial_network}
\end{figure}

Unlike the above-mentioned deep learning-based algorithms, our proposed method not only focuses on the quality of the reconstructed images but also pays attention to the computational resource consumption of the network. We carefully design our network to achieve the best possible balance between the image quality and resource consumption to create a sustainable CS imaging system.

\section{Proposed Method}
In this section, we discuss in detail the introduced RRFM, the proposed compression strategy and the overall architecture of the proposed CSRN. As shown in Fig. \ref{network_structure}, the proposed network consists of three parts: a sampling sub-network, initial reconstruction sub-network and residual reconstruction sub-network, which perform sampling, initial reconstruction and residual reconstruction, respectively. We use $ Conv(n, k, s) $ to denote a convolutional layer and $ PixelShuffle(u) $ to denote the pixel shuffle operation \cite{7780576} for upsampling, where $ n $ is the number of kernels, $ k $ is the kernel size, $ s $ is the stride and $ u $ is the upsampling factor.

\subsection{Sampling Sub-network}
For CS tasks, the sampling step should be as simple as possible. Therefore, in the sampling sub-network, we follow \cite{8765626} and use convolutional layers to sample original images. In traditional block compressed sensing, the image is first divided into small, non-overlapping $ B \times B $ image blocks, and then these small blocks are sampled using the sampling matrix $ \phi $. This process can be mimicked by convolution, as the sampling operation is similar to convolution; each row of the sampling matrix can be replaced by a convolutional layer. For example, given the sample ratio $ r $, the convolutional layer needs $ \lfloor r B^2 \rfloor $ kernels with a kernel size of $ B $ and a stride of $ B $ to sample images. When the sample ratio is 0.1, we use $ Conv(0.1 \times B^2, B, B) $ to sample the original image. Similarly, we use two $ Conv(0.1 \times B^2, B, B) $ layers when the sample ratio is 0.2, and so on. Consequently, the sampling process can be expressed by
\begin{equation}
    \textbf{F}_m = f_s(\textbf{x}),
\end{equation}
where $ f_s(\cdot) $ is the sampling operation, $ \textbf{x} $ is the original image and $ \textbf{F}_m $ represents the measurements. $ \textbf{F}_m $ can be regarded as being composed of $ K $ groups of measurements. Each group of measurements $ \textbf{F}_m^k $ contains $ 0.1 \times B^2 $ feature maps.

Learning the sampling matrix using convolutional layers has two advantages. Compared with a manually designed matrix, on the one hand, the learned matrix takes the inherent characteristics of the signal into account, and on the other hand, the convolution operation guarantees that more structural information can be preserved during the sampling process. Therefore, sampling using convolutional layers guarantees a better reconstruction performance.

\subsection{Initial Reconstruction Sub-network}

According to previous studies, once measurements are obtained, a simple linear mapping can basically recover the original images [34]. We consider this discovery in designing our model and use an initial reconstruction sub-network to obtain initial reconstructed images. However, instead of using a single convolutional layer or fully connected layer to recover images, we use multiple convolutional layers to gradually increase the resolution of feature maps to recover images; we refer to the former method as simple initial reconstruction, while the latter method is referred to as progressive reconstruction. 

Our initial reconstruction sub-network has a hierarchical structure. In other words, the structure of the initial reconstruction sub-network can be different if the corresponding sample ratio is different. When the sample ratio is below 0.1 (for example, if it is 0.05 or 0.01), the sub-network will use a basic recovery block to reconstruct images from process measurements. The basic recovery block contains $ Conv(8m,1,1) $, $ PixelShuffle(4) $, $ Conv(m/2, 3, 1) $, $ Conv(16m,1,1) $ and $ PixelShuffle(8) $, where $ m $ denotes the number of convolution filters and the size of m is equal to B. The 1*1 convolution is designed to explore the information inside the block to recover images, and after the first 1*1 convolution and upsampling operation, the resolution of the feature maps is increased to one eighth of the resolution of the original image. Then, 3*3 convolution is used to extract the information between blocks to mitigate the problem of block artifacts. Then, the second 1*1 convolution and upsampling operation increases the resolution of the feature maps to the resolution of the original image. Finally, $ Conv(1, 3, 1) $ is used to generate the initial reconstructed image. This process can be described as follows:
\begin{equation}
    \textbf{F}_i = f_r(\textbf{F}_m),
\end{equation}
\begin{equation}
    \textbf{x}_i = f_i(\textbf{F}_i),
\end{equation}
where $ f_r(\cdot) $ is used to obtain feature maps with the same resolution as the original images, $ \textbf{F}_i $ represents intermediate feature maps with the resolution of the original images and $f_i(\cdot)$ is the operation for reconstructing the initial image $\textbf{x}_i$.

\begin{algorithm}[t]
	\caption{CSRN} 
	\begin{algorithmic}[1]
	    \Require The original image $\textbf{x}$
	    \Ensure The reconstructed image $\textbf{x}_f$
	    \State /*Sampling the original image $\textbf{x}$ with the sampling sub-network*/
	    \State $\textbf{F}_m \leftarrow f_s(\textbf{x})$
	    \State /*Recovering the initial reconstructed image $\textbf{x}_i$ with the initial reconstruction sub-network*/
	    \For {$k=1,2,\ldots,K$}
	        \State $\textbf{F}_i^k \leftarrow f_r(\textbf{F}_m^k)$
            \If{$k \neq 1$}
                \State $\textbf{F}_i^k \leftarrow f_c([\textbf{F}_i^{k-1},\textbf{F}_i^{k}])$
            \EndIf
        \EndFor
        \State $\textbf{x}_i \leftarrow f_i(\textbf{F}_i^N)$
        \State /*Recovering the residual image $\textbf{x}_r$ with the residual reconstruction sub-network*/
        \State $\textbf{F}_c \leftarrow f_c(\textbf{F}_i^N)$
        \State $\textbf{F}_h \leftarrow f_e(\textbf{F}_c)$
        \State $\textbf{x}_r \leftarrow f_m(\textbf{F}_c,\textbf{F}_h)$
        \State /*Adding the residual image and initial image to obtain the final reconstructed image $\textbf{x}_f$*/
        \State $\textbf{x}_f=\textbf{x}_i+\textbf{x}_r$
	\end{algorithmic} 
\end{algorithm}

When the sample ratio is greater than 0.1, the sub-network will stack basic recovery blocks to process measurement groups and generate intermediate feature maps $ \textbf{F}_i^k $. A series of concatenation and convolution operations are then applied to fuse these intermediate feature maps. Finally, $ Conv(1, 3, 1) $ is used to generate an initial image. For example, when the sample ratio is 0.2, $ Conv(m/2, 1, 1) $ is used to fuse $ \textbf{F}_i^1 $ and $ \textbf{F}_i^2 $, and $ Conv(1, 3, 1) $ is used to generate the initial image. The details of this operation are given below: 
\begin{equation}
    \textbf{F}_i^2 = f_c([\textbf{F}_i^1, \textbf{F}_i^2]),
\end{equation}
\begin{equation}
    \textbf{x}_i = f_i(\textbf{F}_i^2),
\end{equation}
where $ [\cdot] $ denotes the concatenation operation and $ f_c $ is the fuse operation performed using 1*1 convolution. A sketch describing the structure of the initial reconstruction sub-network when the sample ratio is greater than 0.1 is shown in Fig. \ref{initial_network}.

\subsection{Residual Reconstruction Sub-network}

Following the generation of the initially reconstructed image, a residual reconstruction sub-network is applied to generate the residual image in order to preserve the high-frequency details. The residual reconstruction sub-network includes a feature compression module (FCM), a feature extraction module (FEM) with N RRFMs and a feature fusion module (FFM).

As described before, the FCM is located at the head of the residual reconstruction network, and the feature map $\textbf{F}_i^k $ is used as the input of the FCM; the feature map $ \textbf{F}_i^k $ is compressed to obtain the feature map $ \textbf{F}_c $, which has a reduced size:

\begin{equation}
    \textbf{F}_c = f_{FCM}(\textbf{F}_i^k),
\end{equation}
where $ f_{FCM}(\cdot) $ denotes the feature compression operation, and $ \textbf{F}_c $  represents the compressed feature maps. The FCM is described in detail in subsection E. 

The FEM is in the middle of the sub-network, and it stacks $ N $ RRFMs to extract features from $ \textbf{F}_c $:
\begin{equation}
    \textbf{F}_h = f_{FEM}(\textbf{F}_c).
\end{equation}
$ \textbf{F}_h  $ is the output of the FEM. The specific implementation of the RRFM will be presented in subsection D. 

The FFM is located at the end of the sub-network. It can upsample feature maps to recover the original resolution and it fuses feature maps to obtain the residual image. First, the FFM adds $ \textbf{F}_c $ to $ \textbf{F}_h $; then, it uses $ Conv(m, 3, 1) $ and the ReLU activation function to further process the feature maps. Following that, $ PixelShuffle(2) $ is used to upsample the feature maps. In the end, $ Conv(1, 3, 1) $ is used to generate the residual image:
\begin{equation}
    \textbf{x}_r = f_{FFM}(\textbf{F}_c + \textbf{F}_h),
\end{equation}
where $ \textbf{x}_r $ is the residual image and $ f_{FFM} $ is the FFM operation. The final reconstructed image can be generated by adding the residual image and the initial reconstructed image:
\begin{equation}
    \textbf{x}_f = \textbf{x}_i + \textbf{x}_r.
\end{equation}
The pseudocode of CSRN is shown in Algorithm 1.

\subsection{Recurrent Residual Fusion Module}

\begin{figure}
    \centering
    \includegraphics[width=3.0in]{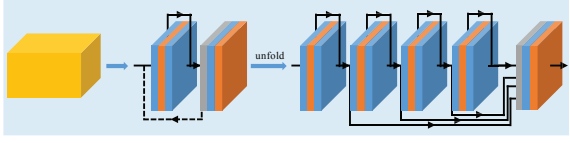}
    \caption{Recurrent residual fusion module. The dotted line denotes the recurrent connection.}
    \label{RRFM}
\end{figure}

The proposed recurrent residual fusion module (RRFM) is composed of a residual block and a convolutional layer. When feature maps flow through the RRFM, the residual block will first recurrently process these features, which means that the output of the residual block from the previous recurrence will be the input of the next recurrence. As the recurrence goes on, the receptive field gradually increases, and thus information about the object from different scales can be extracted. Then, the multi-scale intermediate features obtained by each recurrence will be concatenated. Finally, the convolutional layer will fuse these concatenated feature maps; the fused feature maps are the output of the RRFM.

The residual block was proposed in \cite{7780459} to facilitate gradient propagation; it consists of a convolutional layer, batch normalization (BN) layer and activation function. It has shown great success in many computer tasks. However, the original design of the residual block is high-level-vision-task-oriented and thus not suitable for low-level vision tasks due to the fact that the BN layer causes the network to lose flexibility through feature normalization. Therefore, EDSR \cite{8014885} removes the BN layer in the residual block and thus achieves a better performance in the super-resolution area. Since image compressive sensing is essentially a low-level vision task, we adopt this improved residual block in our network. The convolutional layer in the RRFM accepts intermediate feature maps generated by the residual block during recurrence and performs feature fusion. Therefore, we set both the kernel size and the stride of the convolutional layer to 1, that is, $ Conv(m,1,1) $ is used.

As shown in Fig. \ref{RRFM}, due to the recurrent structure of the RRFM, it can be unfolded into T residual blocks with shared parameters and a convolutional layer. The output $ \textbf{F}_r $ of the RRFM is
\begin{equation}
    \textbf{F}_r = f_{fuse}([\textbf{R}^1, \textbf{R}^2,\dots, \textbf{R}^t]),
\end{equation}
where $ t=1,2,…,T $, $ \textbf{R}^t $ represents the intermediate features generated by the $n$th recurrence, $ [\cdot] $ denotes the concatenation operation and $ f_{fuse} (\cdot) $ denotes the feature fusion operation performed using $ Conv(m, 1, 1) $.

\begin{figure}[t]
    \centering
    \includegraphics[width=3.0in]{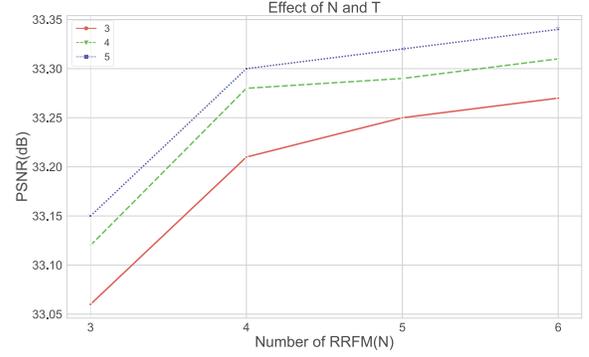}
    \caption{Reconstruction performance on Set5 for different combinations of N and T.}
    \label{nt_test}
\end{figure}

\subsection{Feature Compression Strategy}
To speed up and reduce the memory consumption of the image reconstruction process, we introduce a feature compression strategy into the residual reconstruction sub-network. This strategy compresses the size of feature maps. We justify the introduction of such a compression strategy using the following argument. Since the initial reconstruction sub-network can reconstruct a relatively good initial image, the residual module does not need to retain the feature maps with the original resolution of the image to obtain its high-frequency details. Therefore, the feature compression module (FCM) is placed at the head of the residual reconstruction sub-network to perform the compression operation. This block takes feature maps from the initial reconstruction sub-network as the input and compresses the feature maps to a quarter of their original size; then, it passes the compressed feature maps to the rest of the residual reconstruction network for further processing. 

We use $ Conv(m, 2, 2) $ and the ReLU activation function to form the FCM. The stride convolution layer can compress feature maps to one quarter of their original size, and thus the benefits of compressing feature maps are obvious. First, the compressed feature maps take up less memory, so a deeper network can be constructed for a better performance without having to consider the memory consumption. Second, the number of calculations will be reduced, along with the burden placed on the computing devices being used.

\section{Experimental Studies}

In this section, we will introduce the settings and datasets used in the experiments. Our proposed network will be compared with representative existing methods, and the contributions of each individual part of our proposed network will also be analyzed and discussed.

\subsection{Settings and Datasets}

To evaluate our proposed network, we use 7 sample ratios, including 0.01, 0.05, 0.1, 0.2, 0.3, 0.4 and 0.5; we use 400 images from BSDS500 \cite{9019857} as the training set and 100 images as the validation set. Each image of the training set is cropped into small $96 \times 96$ pieces. Eight data augmentation methods described in \cite{7780551} are further applied to make full use of these images. As a result, we have in total 89600 image pieces for training and 22400 image pieces for validation. The Adam optimizer is used to train CSRN with a batch size of 16 for 100 epochs. The learning rate starts at 0.0005 and decreases by a factor of 0.1 at the 50\textsuperscript{th} and 80\textsuperscript{th} epochs. Higher learning rates at the beginning of training are more likely to suppress the memory of noisy data and speed up training, while lower learning rates in the middle and later stages of training make it easier to find the global optimal solution. The number of convolution filters $ m $ is set to 32. We use the mean square error(MSE) loss of the initial reconstructed image and final reconstructed image to guide the gradient descent process: 
\begin{equation}
\begin{split}
    \Theta( \Lambda ) = \underset{\bm{\theta}_s, \bm{\theta}_i, \bm{\theta}_r}{\arg\min}\frac{1}{2M}\sum_{i=1}^M \norm{g(\textbf{x}^i;\bm{\theta}_s, \bm{\theta}_i) - \textbf{x}^i}_2^2 + \\
    \frac{1}{2M}\sum_{i=1}^M \norm{g(\textbf{x}^i; \bm{\theta}_s, \bm{\theta}_i, \bm{\theta}_r) - \textbf{x}^i}_2^2,
\end{split}
\end{equation}
where $ \Lambda $ denotes the collection
of model parameters, $ \bm{\theta}_s $, $ \bm{\theta}_i $ and $ \bm{\theta}_r $ denote the parameters of the sampling sub-network, initial reconstruction sub-network and residual reconstruction sub-network, respectively, $ g(\cdot) $ represents the network's operation, $ g(\textbf{x}^i;\bm{\theta}_s, \bm{\theta}_i) $ denotes the initial reconstructed image and $ g(\textbf{x}^i;\bm{\theta}_s, \bm{\theta}_i, \bm{\theta}_r) $ denotes the final reconstructed image. We take the model that obtains the smallest loss on the validation set as our final model and test its performance. We implement the proposed network in PyTorch, and we train and test our models on an NVIDIA Titan Xp.

The Set5 \cite{bevilacqua:hal-00747054}, Set11 \cite{7780424}, Set14 \cite{10.1007/978-3-642-27413-8_47}, BSD100 \cite{937655}, Urban100 \cite{7299156} and DIV2K validation sets \cite{8014883} are used as test datasets in our experiments. We measure the reconstruction quality using the peak signal-to-noise ratio (PSNR) \cite{hore2010image} and structural similarity index (SSIM)  \cite{wang2004image}. Like other CS algorithms \cite{zhao2019roi,yamac2021generalized}, we convert the image from the RGB space into the YCbCr space and take the luminance channel as the input. 

\subsection{Determining the Values of N and T}

To choose the best combination of $ N $ and $ T $, we combine different numbers of RRFMs $ (N) $ with different numbers of recurrences $ (T) $ in the residual reconstruction sub-network for a sample ratio of 0.1 and test the performance of this sub-network on Set5.

The results are shown in Fig. \ref{nt_test}; we can see that using more RRFMs and increasing the number of recurrences leads to a better reconstruction quality. Stacking more RRFMs and making the network deeper make it possible to extract more meaningful features for high-frequency details and hence enrich the reconstructions. Furthermore, when $ N $ is large, increasing $ T $ can still improve the reconstruction quality, but the improvement is not as obvious as when $ N $ is small. Although increasing $ N $ and $ T $ could provide better performances, the network will use more memory and time for reconstruction. Hence, after weighing the performance of the network against the resources required by  the network, we set $ N $ to 5 and $ T $ to 3 in our experiments.

\subsection{Comparison with State-of-the-Art Methods}
\begin{table*}[t]
	\caption{\centering Reconstruction Quality Comparison of CSRN, Optimization-Based Algorithms and Plain Networks}
	\begin{center}
		\renewcommand\arraystretch{1}
        
		\setlength{\tabcolsep}{0.6mm}{
			\begin{tabular}{ccccccccc}
				\hline
				\hline
				\multirow{2}{*}{Algorithm}
				& 0.01 & 0.05 & 0.1 & 0.2 & 0.3 & 0.4 & 0.5 & Average\\
				\cline{2-9}
				& PSNR(dB) \ SSIM & PSNR(dB) \ SSIM & PSNR(dB) \ SSIM & PSNR(dB) \ SSIM & PSNR(dB) \ SSIM & PSNR(dB) \ SSIM & PSNR(dB) \ SSIM & PSNR(dB) \ SSIM\\
				\hline	
				\multicolumn{9}{c}{Set5}\\
				\hline
				DWT 			  & 13.76/0.4419 & 23.11/0.6831 & 26.58/0.7601 & 29.74/0.8366 & 31.98/0.8790 & 33.81/0.9077 & 35.53/0.9299 & 27.79/0.7769\\
				MH 			  & 15.38/0.3718 & 25.25/0.7192 & 28.57/0.8211 & 31.97/0.8859 & 33.73/0.9123 & 35.27/0.9324 & 36.86/0.9474 & 29.57/0.7986\\
				GSR 		  & 21.03/0.5882 & 26.97/0.7944 & 30.48/0.8754 & 34.54/0.9286 & 36.94/0.9498 & 38.85/0.9627 & 40.65/0.9724 & 32.78/0.8674\\
				ReconNet	     & 20.34/0.5347 & 25.26/0.7039 & 28.03/0.7943 & 30.77/0.8597 & 32.36/0.8905 & 34.21/0.9172 & 34.92/0.9273 & 29.41/0.8039\\
				Dr\textsuperscript{2}-Net     &  20.37/0.5384 & 25.56/0.7205 & 28.50/0.8180 & 31.49/0.8813 & 33.71/0.9178 & 35.40/0.9366 & 37.08/0.9522 & 30.30/0.8235\\
				DPA-Net & 19.02/0.5133 & 26.66/0.7893 & 29.29/0.8705 & 32.54/0.9019 & 34.57/0.9412 & 36.13/0.9575 & 37.51/0.9679 & 30.82/0.8488 \\ 
				SCSNet & 24.21/0.6468 & 29.74/0.8472 & 32.77/0.9083 & 36.15/0.9487 & 38.45/ 0.9655 & 40.44/0.9755 & 42.22/0.9820 & 34.85/0.8963\\
				CSNet\textsuperscript{+}        & 24.23/0.6492 & 29.80/0.8506 & 32.65/0.9078 & 36.21/0.9489 & 38.53/0.9656 & 40.32/0.9750 & 42.27/0.9819 & 34.85/0.8970\\	
				CSRN   & \textbf{24.53/0.6628} & \textbf{30.39/0.8638} & \textbf{33.25/0.9155} & \textbf{36.61/0.9518} & \textbf{38.83/0.9673} & \textbf{40.72/0.9765} & \textbf{42.59/0.9830} & \textbf{35.27/0.9030}\\
				\hline
				\multicolumn{9}{c}{Set11}\\
				\hline			
				DWT 			  & 13.24/0.3731 & 21.23/0.5980 & 23.41/0.6852 & 26.32/0.7840 & 28.43/0.8407 & 30.34/0.8806 & 32.14/0.9105 & 25.02/0.7246\\
				MH 			  & 14.11/0.3374 & 22.24/0.6545 & 26.07/0.7873 & 29.39/0.8671 & 31.51/0.9039 & 33.29/0.9266 & 34.90/0.9438 & 27.36/0.7744\\
				GSR 		  & 18.19/0.4874 & 24.05/0.7516 & 28.12/0.8585 & 32.25/0.9216 & 34.77/0.9461 & 36.87/0.9614 & 38.71/0.9716 & 30.42/0.8426\\
				ReconNet	     & 17.83/0.4541 & 22.07/0.6276 & 24.65/0.7345 & 27.30/0.8170 & 28.89/0.8546 & 30.90/0.8941 & 31.60/0.9064 & 26.18/0.7555\\
				Dr\textsuperscript{2}-Net     &  17.87/0.4595 & 22.40/0.6449 & 25.16/0.7568 & 28.12/0.8399 & 30.33/0.8845 & 32.06/0.9114 & 33.77/0.9329 & 27.10/0.7757\\
				DPA-Net & 18.20/0.5101 & 23.77/0.7811 & 27.66/0.8530 & 30.47/0.9087 & 33.35/0.9425 & 35.21/0.9580 & 36.80/0.9685 & 29.35/0.8460\\ 
				SCSNet &  21.04/0.5601 & 25.79/0.7878 & 28.48/0.8630 & 31.97/0.9229 & 34.62/0.9512 &  36.92/0.9652 & 39.01/0.9767 & 31.12/0.8610 \\
				CSNet\textsuperscript{+}        & 21.02/0.5586 & 25.98/0.7891 & 28.54/0.8632 & 32.02/0.9230 & 34.70/0.9511 & 36.80/0.9657 & 39.03/0.9768 & 31.16/0.8611\\	
				CSRN   & \textbf{21.40/0.5751} & \textbf{26.57/0.8076} & \textbf{29.00/0.8733} & \textbf{32.55/0.9293} & \textbf{35.21/0.9556} & \textbf{37.49/0.9699} & \textbf{39.60/0.9790} & \textbf{31.69/0.8700}\\
				\hline
                \multicolumn{9}{c}{Set14}\\
				\hline			
				DWT 			  & 14.83/0.4097 & 22.44/0.5939 & 24.73/0.6684 & 27.43/0.7584 & 29.32/0.8159 & 31.00/0.8589 & 32.63/0.8925 & 26.05/0.7140\\
				MH 			  & 15.16/0.3517 & 23.47/0.6156 & 26.38/0.7433 & 29.48/0.8228 & 31.29/0.8674 & 32.89/0.8989 & 34.43/0.9225 & 27.59/0.7460\\
				GSR 		  & 20.32/0.5124 & 24.67/0.6756 & 27.87/0.7775 & 31.40/0.8622 & 33.80/0.9046 & 35.82/0.9322 & 37.64/0.9513 & 30.22/0.8022\\
				ReconNet	     & 19.44/0.4639 & 23.52/0.6033 & 25.64/0.6916 & 27.78/0.7689 & 29.13/0.8135 & 30.86/0.8594 & 31.45/0.8751 & 26.83/0.7251\\
				Dr\textsuperscript{2}-Net     &  19.48/0.4683 & 23.79/0.6154 & 25.94/0.7066 & 28.33/0.7912 & 30.40/0.8481 & 31.87/0.8835 & 33.35/0.9096 & 27.59/0.7461\\
				DPA-Net & 18.30/0.4616 & 24.71/0.7286 & 26.28/0.7693 & 31.47/0.8845 & 30.86/0.8978 & 32.30/0.9246 & 33.78/0.9440 & 28.24/0.8015\\
				SCSNet & 22.87/0.5631 & 26.92/0.7322 & 29.22/0.8181 & 32.19/0.8945 & 34.51/ 0.9311 & 36.54/0.9525 & 38.41/0.9659 & 31.52/0.8368\\
				CSNet\textsuperscript{+}        & 22.77/0.5577 & 26.88/0.7313 & 29.11/0.8167 & 32.19/0.8942 & 34.52/0.9306 & 36.41/0.9514 & 38.39/0.9656 & 31.47/0.8354\\	
				CSRN   & \textbf{23.04/0.5722} & \textbf{27.23/0.7428} & \textbf{29.43/0.8251} & \textbf{32.48/0.8993} & \textbf{34.75/0.9340} & \textbf{36.73/0.9546} & \textbf{38.72/0.9674} & \textbf{31.77/0.8422}\\
				\hline
                \multicolumn{9}{c}{BSD100}\\
				\hline			
				DWT 			  & 15.39/0.4135 & 23.12/0.5765 & 24.98/0.6449 & 27.14/0.7332 & 28.75/0.7938 & 30.25/0.8407 & 31.70/0.8781 & 25.90/0.6972\\
				MH 			  & 16.41/0.3730 & 23.67/0.5865 & 25.16/0.6673 & 28.11/0.7733 & 29.74/0.8287 & 31.22/0.8688 & 32.65/0.8998 & 26.71/0.7139\\
				GSR 		  & 21.14/0.4997 & 23.95/0.6204 & 25.98/0.7094 & 28.79/0.8075 & 30.98/0.8662 & 32.93/0.9059 & 34.77/0.9340 & 28.36/0.7633\\
				ReconNet	     & 20.87/0.4761 & 24.08/0.5905 & 25.74/0.6698 & 27.49/0.7434 & 28.66/0.7904 & 30.29/0.8450 & 30.91/0.8641 & 26.86/0.7113\\
				Dr\textsuperscript{2}-Net     &  20.92/0.4808 & 24.25/0.5984 & 25.98/0.6815 & 28.07/0.7689 & 29.80/0.8308 & 31.26/0.8731 & 32.75/0.9045 & 27.57/0.7340\\
				DPA-Net & 19.00/0.4487 & 23.51/0.6771 & 25.37/0.7219 & 27.78/0.8413 & 29.55/0.8735 & 31.05/0.9089 & 32.44/0.9341 & 26.96/0.7722\\ 
				SCSNet & 23.78/0.5483 & 26.77/0.6972 & 28.57/0.7844 & 31.10/0.8731 & 33.24/ 0.9190 & 35.21/0.9470 & 37.14/0.9649 & 30.83/0.8191\\
				CSNet\textsuperscript{+}        & 23.76/0.5478 & 26.81/0.6990 & 28.60/0.7847 & 31.13/0.8730 & 33.29/0.9194 & 35.21/0.9467 & 37.17/0.9649 & 30.85/0.8194\\	
				CSRN   & \textbf{23.98/0.5564} & \textbf{27.01/0.7067} & \textbf{28.79/0.7916} & \textbf{31.35/0.8777} & \textbf{33.45/0.9220} & \textbf{35.39/0.9487} & \textbf{37.33/0.9661} & \textbf{31.04/0.8242}\\
                \hline
				\multicolumn{9}{c}{Urban100}\\
				\hline
				ReconNet	     & 17.94/0.4303 & 21.26/0.5705 & 23.23/0.6670 & 25.28/0.7523 & 26.50/0.7958 & 28.30/0.8481 & 28.79/0.8624 & 24.47/0.7038\\
				Dr\textsuperscript{2}-Net     &  17.97/0.4351 & 21.54/0.5876 & 23.73/0.6951 & 26.17/0.7884 & 28.21/0.8471 & 29.70/0.8810 & 31.29/0.9097 & 25.52/0.7349\\
				DPA-Net & 16.36/0.4162 & 22.88/0.6404 & 24.55/0.7851 & 28.82/0.8093 & 29.48/0.9044 & 31.10/0.9319 & 32.09/0.9454 & 26.47/0.7761\\ 
				SCSNet & 20.65/0.5196 & 24.37/0.7171 & 26.61/0.8093 & 29.60/0.8915 & 32.01/0.9310 & 34.07/0.9531 & 36.10/0.9680 & 29.06/0.8271\\
				CSNet\textsuperscript{+}        & 20.69/0.5199 & 24.38/0.7171 & 26.63/0.8098 & 29.65/0.8917 & 32.04/0.9311 & 34.12/0.9533 & 36.12/0.9681 & 29.09/0.8273\\	
				CSRN   & \textbf{20.89/0.5314} & \textbf{24.74/0.7336} & \textbf{26.96/0.8225} & \textbf{30.09/0.9005} & \textbf{32.43/0.9359} & 34.52/\textbf{0.9570} & 36.43/\textbf{0.9704} & \textbf{29.44/0.8359}\\
				\hline
				\multicolumn{9}{c}{DIV2K validation set}\\
				\hline			
				ReconNet	     & 21.81/0.5862 & 25.81/0.7001 & 27.89/0.7704 & 29.99/0.8301 & 31.38/0.8644 & 33.16/0.9004 & 33.79/0.9124 & 29.12/0.7949\\
				Dr\textsuperscript{2}-Net     &  21.87/0.5922 & 26.07/0.7113 & 28.24/0.7842 & 30.73/0.8508 & 32.72/0.8938 & 34.32/0.9209 & 35.92/0.9407 & 29.98/0.8134\\
				DPA-Net & 19.79/0.5508 & 26.81/0.7648 & 28.21/0.8219 & 31.69/0.8711 & 33.02/0.9197 & 34.49/0.9420 & 35.97/0.9577 & 30.00/0.8326\\ 
				SCSNet & 25.17/0.6631 & 29.10/0.8010 & 31.35/0.8673 & 34.34/0.9251 & 36.71/0.9525 & 38.88/0.9688 & 40.41/0.9699 & 33.71/0.8782\\
				CSNet\textsuperscript{+}        & 25.15/0.6630 & 29.11/0.8010 & 31.34/0.8671 & 34.36/0.9254 & 36.73/0.9528 & 38.86/0.9686 & 40.86/0.9786 & 33.77/0.8795\\	
				CSRN   & \textbf{25.41/0.6708} & \textbf{29.42/0.8090} & \textbf{31.65/0.8731} & \textbf{34.69/0.9293} & \textbf{37.05/0.9553} & \textbf{39.15/0.9703} & \textbf{41.16/0.9798} & \textbf{34.08/0.8839}\\
				\hline
				\hline
		\end{tabular}}
		\label{quality1}
	\end{center}
\end{table*}

\begin{table*}[t]
	\caption{\centering Reconstruction Quality Comparison of CSRN and Deep Unfolding Networks}
	\begin{center}
		\renewcommand\arraystretch{1}
		\setlength{\tabcolsep}{1.5mm}{
			\begin{tabular}{ccccccc}
				\hline
				\hline
				\multirow{2}{*}{Algorithm}
				& 0.1 & 0.2 & 0.3 & 0.4 & 0.5 & Average\\
				\cline{2-7}
				& PSNR(dB) \ SSIM & PSNR(dB) \ SSIM & PSNR(dB) \ SSIM & PSNR(dB) \ SSIM & PSNR(dB) \ SSIM & PSNR(dB) \ SSIM\\
				\hline	
				\multicolumn{7}{c}{Set5}\\
				\hline
				ISTA-Net\textsuperscript{+}   & 30.31/0.8627 & 34.37/0.9218 & 36.90/0.9471 & 38.91/0.9611 & 40.71/0.9733 & 36.24/0.9332\\
                ISTA-Net\textsuperscript{+}\textsuperscript{+}   & 30.22/0.8702 & 33.94/0.9250 & 36.27/0.9485 & 38.13/0.9621 & 39.95/0.9725 & 35.70/0.9357\\
                COAST & 30.50/0.8794 & 34.18/0.9298 & 36.48/0.9515 & 38.33/0.9645 & 40.21/0.9744 & 35.94/0.9399\\
				OPINE-Net & 32.51/0.9150 & 35.55/0.9471 & 37.44/0.9579 & 39.84/0.9648 & 41.62/0.9805 & 37.39/0.9531\\
				CSRN  & \textbf{33.25/0.9155} & \textbf{36.61/0.9518} & \textbf{38.83/0.9673} & \textbf{40.72/0.9765} & \textbf{42.59/0.9830} & \textbf{38.40/0.9588}\\
				\hline
				\multicolumn{7}{c}{Set11}\\
				\hline			
				ISTA-Net\textsuperscript{+}  & 27.02/0.8135 & 31.12/0.9002 & 34.00/0.9359 & 36.17/0.9549 & 38.07/0.9706 & 33.28/0.9150\\
                ISTA-Net\textsuperscript{+}\textsuperscript{+}   & 28.34/0.8531 & 32.33/0.9217 & 34.86/0.9478 & 36.94/0.9628 & 38.73/0.9727 & 34.24/0.9316\\
                COAST & 28.69/0.8618 & 32.54/0.9251 & 35.04/0.9501 & 37.13/0.9648 & 38.94/0.9744 & 34.47/0.9352\\
				OPINE-Net & \textbf{29.81/0.8904} & 32.51/0.9291 & 34.57/0.9328 & 36.17/0.9533 & \textbf{40.19/0.9800} & 34.65/0.9371\\
				CSRN   & 29.00/0.8733 & \textbf{32.55/0.9293} & \textbf{35.21/0.9556} & \textbf{37.49/0.9699} & 39.60/0.9790 & \textbf{34.77/0.9414}\\
				\hline
                \multicolumn{7}{c}{Set14}\\
				\hline			
				ISTA-Net\textsuperscript{+}   & 27.31/0.7478 & 30.86/0.8476 & 33.45/0.8959 & 35.51/0.9276 & 37.08/0.9525 & 32.84/0.8743\\
                ISTA-Net\textsuperscript{+}\textsuperscript{+}   & 27.32/0.7714 & 30.72/0.8628 & 33.07/0.9074 & 34.98/0.9341 & 36.73/0.9525 & 32.56/0.8856\\
                COAST & 27.41/0.7799 & 30.71/0.8672 & 33.10/0.9106 & 35.12/0.9369 & 36.94/0.9549 & 32.66/0.8899\\
				OPINE-Net & 28.77/\textbf{0.8294} & 31.82/0.8847 & 33.59/0.9315 & 36.19/0.9424 & 38.09/0.9662 & 33.69/0.9108\\
				CSRN   & \textbf{29.43}/0.8251 & \textbf{32.48/0.8993} & \textbf{34.75/0.9340} & \textbf{36.73/0.9546} & \textbf{38.72/0.9674} & \textbf{34.42/0.9161}\\
				\hline
                \multicolumn{7}{c}{BSD100}\\
				\hline			
				ISTA-Net\textsuperscript{+}   & 26.79/0.7138 & 29.54/0.8161 & 31.67/0.8751 & 33.53/0.9130 & 35.05/0.9404 & 31.32/0.8517\\
                ISTA-Net\textsuperscript{+}\textsuperscript{+}   & 26.15/0.7261 & 28.85/0.8301 & 30.91/0.8866 & 32.80/0.9225 & 34.63/0.9470 & 30.67/0.8625\\
                COAST & 26.18/0.7297 & 28.83/0.8302 & 30.86/0.8851 & 32.71/0.9204 & 34.49/0.9450 & 30.61/0.8621\\
				OPINE-Net & 27.55/0.7903 & 29.58/0.8526 & 32.17/0.8942 & 33.98/0.9382 & 36.02/0.9627 & 31.86/0.8876\\
				CSRN   & \textbf{28.79/0.7916} & \textbf{31.35/0.8777} & \textbf{33.45/0.9220} & \textbf{35.39/0.9487} & \textbf{37.33/0.9661} & \textbf{33.26/0.9012}\\
                \hline
				\multicolumn{7}{c}{Urban100}\\
				\hline
				ISTA-Net\textsuperscript{+}   & 25.38/0.7572 & 29.42/0.8703 & 32.29/0.9195 & \textbf{34.59}/0.9466 & \textbf{36.74}/0.9641 & 31.68/0.8915\\
                ISTA-Net\textsuperscript{+}\textsuperscript{+}   & 25.53/0.7882 & 29.36/0.8867 & 31.93/0.9280 & 33.99/0.9504 & 35.84/0.9646 & 31.33/0.9036\\
                COAST & 25.94/0.8038 & 29.70/0.8940 & 32.20/0.9317 & 34.21/0.9528 & 35.99/0.9665 & 31.61/0.9098\\	
				OPINE-Net & 26.61/\textbf{0.8362} & 30.01/0.8942 & 32.38/0.9318 & 34.19/0.9505 & 36.62/\textbf{0.9727} & 31.96/0.9171\\
			
				CSRN   & \textbf{26.96}/0.8225 & \textbf{30.09/0.9005} & \textbf{32.43/0.9359} & 34.52/\textbf{0.9570} & 36.43/0.9704 & \textbf{32.09/0.9173}\\
				\hline
				\multicolumn{7}{c}{DIV2K validation set}\\
				\hline			
				ISTA-Net\textsuperscript{+}   & 29.40/0.8139 & 32.71/0.8894 & 35.15/0.9276 & 37.19/0.9504 & 39.22/0.9661 & 34.73/0.9095\\
                ISTA-Net\textsuperscript{+}\textsuperscript{+}   & 28.96/0.8209 & 32.11/0.8944 & 34.41/0.9304 & 36.38/0.9517 & 38.22/0.9656 & 34.02/0.9126\\
                COAST & 29.16/0.8279 & 32.22/0.8972 & 34.45/0.9318 & 36.40/0.9528 & 38.22/0.9668 & 34.09/0.9153\\
				OPINE-Net & 30.64/0.8713 & 32.98/0.9184 & 35.63/0.9422 & 38.07/0.9678 & 39.88/0.9767 & 35.44/0.9353\\
				CSRN   & \textbf{31.65/0.8731} & \textbf{34.69/0.9293} & \textbf{37.05/0.9553} & \textbf{39.15/0.9703} & \textbf{41.16/0.9798} & \textbf{36.74/0.9416}\\
				\hline
				\hline
		\end{tabular}}
		\label{quality2}
	\end{center}
\end{table*}


\begin{figure*}[t]
    \centering
    \includegraphics[width=0.95\textwidth]{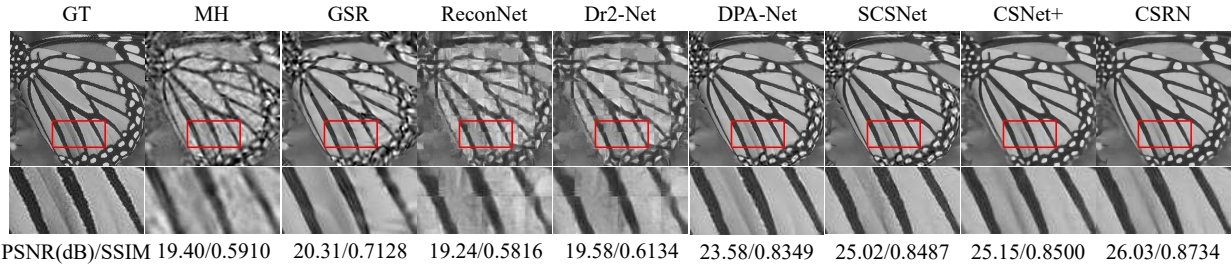}
    \caption{Visual quality comparison of different optimization- and plain-network-based CS algorithms on ``Butterfly" from Set5 for the sample ratio 0.05.}
    \label{placeholder1}
\end{figure*}

\begin{figure*}[t]
    \centering
    \includegraphics[width=0.95\textwidth]{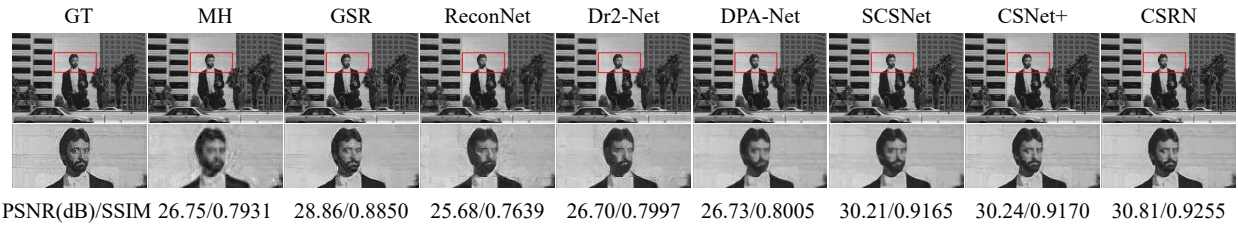}
    \caption{Visual quality comparison of different optimization- and plain-network-based CS algorithms  on ``119082" from BSD100 for the sample ratio 0.2.}
    \label{placeholder3}
\end{figure*}

\begin{figure*}[t]
    \centering
    \includegraphics[width=0.95\textwidth]{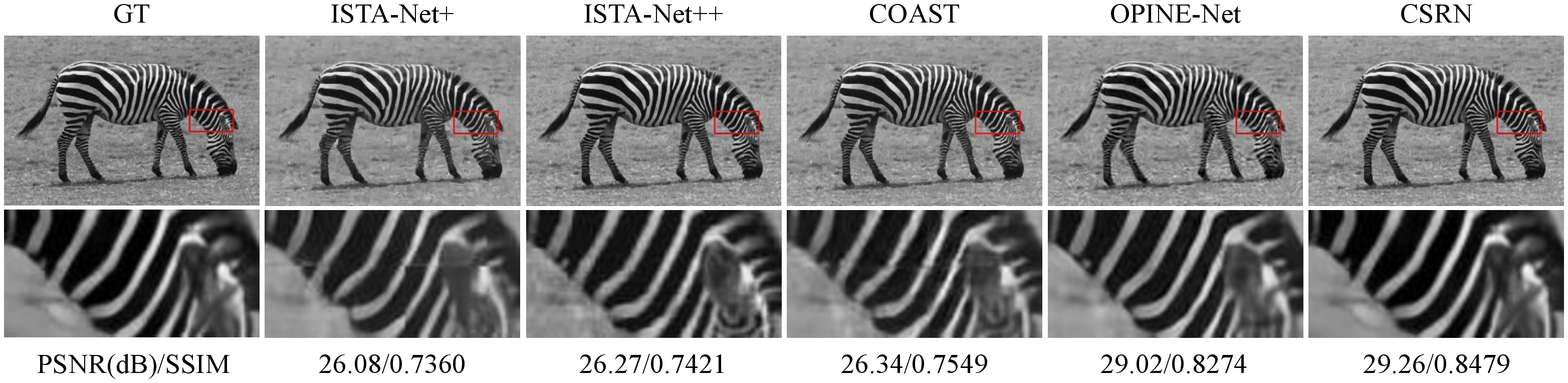}
    \caption{Visual quality comparison of different deep unfolding network-based CS algorithms on ``Zebra" from Set14 for the sample ratio 0.1.}
    \label{placeholder2}
\end{figure*}

To provide a comprehensive evaluation of our proposed method, we compare the proposed method with DWT \cite{5453522}, MH \cite{6190204}, GSR \cite{6814320}, ReconNet \cite{7780424}, Dr\textsuperscript{2}-Net \cite{YAO2019483}, DPA-Net \cite{9199540}, SCSNet \cite{8953841}, CSNet\textsuperscript{+} \cite{8765626}, ISTA-Net\textsuperscript{+} \cite{8578294}, ISTA-Net\textsuperscript{+}\textsuperscript{+} \cite{ 9428249}, OPINE-Net \cite{9019857} and COAST \cite{9467810}; the first three methods are optimization-based methods, the middle five methods are plain networks and the last four methods are deep unfolding networks. For all of these methods, we used the recommended parameter settings and the code implementations provided by the authors. Particularly, since OPINE-Net and COAST apply multiple sampling matrices within one network, for a fair comparison and to focus on the reconstruction ability of our CSRN, we only use a single learned sampling matrix for both of these methods. We calculate the reconstruction quality of deep learning-based methods on all six test datasets, but we test the optimization-based methods only on Set5, Set11, Set14 and BSD100 since optimization-based methods take too much time on the Urban100 and DIV2K validation sets. 

\textbf{Comparison with optimization-based algorithms and plain deep networks:} We first compare CSRN with optimization-based algorithms and plain deep networks. The results are shown in Table \ref{quality1}, where it can be seen that compared with optimization-based methods, deep learning-based methods have a better reconstruction quality in most cases. As DWT and MH use a manual sparse basis to process CS measurements, they have relatively poor performances. GSR merges group sparsity into the reconstruction process and thus is capable of further exploring the image information. ReconNet has a simple network structure, yet its performance is better than that of DWT but worse than that of GSR. Dr\textsuperscript{2}-Net introduces a residual connection into its network structure, and its reconstruction performance is better than that of ReconNet. Furthermore, it is worthwhile to note that ReconNet and Dr\textsuperscript{2}-Net can achieve better results than DWT and MH when sample ratios less than 0.05 are used, which shows the potential of deep learning-based CS algorithms at low sample ratios. DPA-Net, SCSNet and CSNet\textsuperscript{+} have more sophisticated structures, so their reconstruction qualities surpass those of optimization-based methods. CSRN achieves the best average reconstruction quality out of all eight compared methods on all test sets. Compared with the second-best method, CSRN improves the PNSR by 0.42, 0.53, 0.25, 0.19, 0.35 and 0.31 dB on average, and it improves the SSIM by 0.0060, 0.0089, 0.0054, 0.0048, 0.0086 and 0.0044 on average on Set5, Set11, Set14, BSD100, Urban100 and DIV2k, respectively. All the experimental results validate the effectiveness of our proposed CSRN. 

Some reconstructed samples are illustrated in Fig. \ref{placeholder1} and Fig. \ref{placeholder3} for visual inspection, and we can observe that the images reconstructed by optimization-based methods have some noise, which impairs the visual quality of the reconstructed images. As shown, the images reconstructed by MH have severe blurring and artifacts that affect the visual appearance. GSR performs best out of the optimization methods; it works well for smooth areas of the image but still produces serious artifacts in areas with more texture. For plain deep networks, ReconNet and Dr\textsuperscript{2}-Net produce significant blocky artifacts, while DPA-Net, SCSNet and CSNet\textsuperscript{+} present a better visual quality by using a deeper feature extraction structure to retain smooth structures. However, the boundary or intersection between the different structures is usually blurred or over-smoothed, which indicates that some details have been lost. Compared with these methods, our proposed CSRN not only maintains a structural smoothness but also reconstructs more boundary details through multi-scale feature fusion and the recurrent learning process.

\textbf{Comparison with deep unfolding networks:} We further compare CSRN with deep unfolding networks and show the results in Table \ref{quality2}. It can be seen that due to their physics-driven structures, ISTA-Net and OPINE-Net obtain a better reconstruction quality under some circumstances, but CSRN still achieves a better average reconstruction quality compared with deep unfolding algorithms. All of the above experimental results validate the effectiveness of our proposed CSRN.

\begin{figure*}[t]
    \centering
    \includegraphics[width=0.95\textwidth]{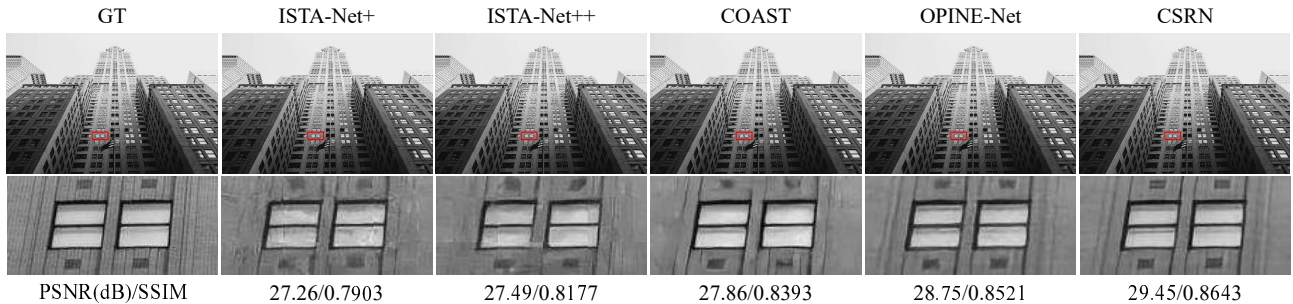}
    \caption{Visual quality comparison of different deep-unfolding-network-based CS algorithms on ``0846" from DIV2K for a sample ratio of 0.1.}
    \label{placeholder4}
\end{figure*}

\begin{figure*}[t]
    \centering
    \includegraphics[width=0.95\textwidth]{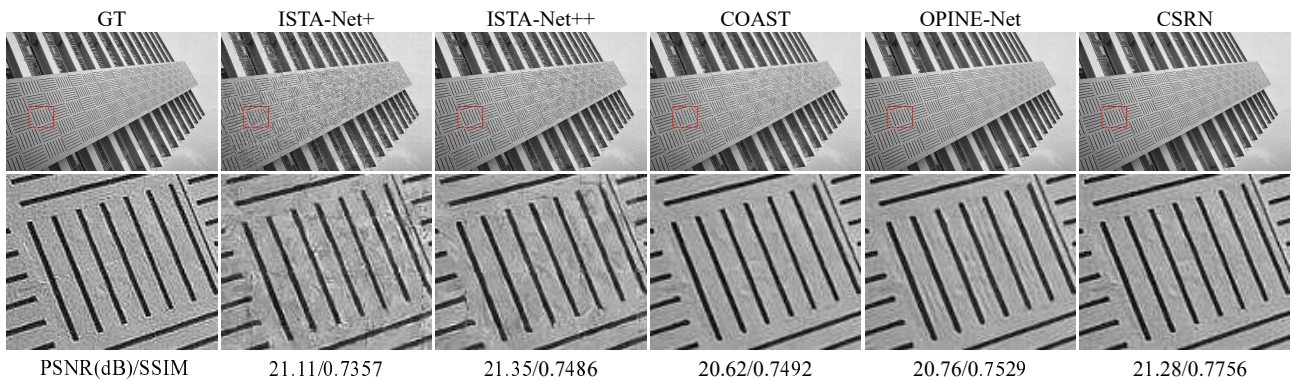}
    \caption{Visual quality comparison of different deep-unfolding-network-based CS algorithms on ``img 092" from Urban100 for a sample ratio of 0.2.}
    \label{placeholder5}
\end{figure*}

The visual results are displayed in Fig. \ref{placeholder2}, Fig. \ref{placeholder4} and Fig. \ref{placeholder5}. ISTA-Net\textsuperscript{+}, ISTA-Net\textsuperscript{++} and COAST perform well in recovering the in-block information of images, but they incur some block artifacts due to the fact that they reconstruct each image block independently. To solve the block artifacts problem, OPINE-Net extends the intra-block training to inter-block training to enhance the CS recovery quality. CSRN can better mitigate these artifacts because the model optimizes the whole image, not just small image patches, during the training stage. Compared with these methods, CSRN recovers images more clearly due to its effective network design. In particular, CSRN retains more details and has better visual effects. Therefore, it is concluded that CSRN has a better performance according to both subjective and objective evaluations. 

\textbf{Reconstruction time comparison:} The average reconstruction times for $256 \times 256$ images achieved by CSRN and the compared algorithms are summarized in Table \ref{time_comparison}. Overall, compared with optimization-based methods, deep learning methods reconstruct images much faster due to their non-iterative nature. The three considered optimization-based methods take roughly a few seconds to several minutes to reconstruct an image. CSRN is significantly faster than most of the deep learning-based algorithms, including DPA-Net, SCSNet, ISTA-Net\textsuperscript{+}, ISTA-Net\textsuperscript{++}, COAST and OPINE-Net. CSRN employs a recurrent structure, which deepens the network, and thus its time is comparable to that of CSNet\textsuperscript{+}. Although ReconNet and Dr\textsuperscript{2}-Net run faster than CSRN, both of them employ a shallow structure that fails to reconstruct the details perfectly, thus resulting in low PSNR and SSIM values compared with CSRN. In short, CSRN can reconstruct images faster than its competitors and maintain a high reconstruction quality.  

\begin{figure}[h]
    \centering
    \includegraphics[width=3.0in]{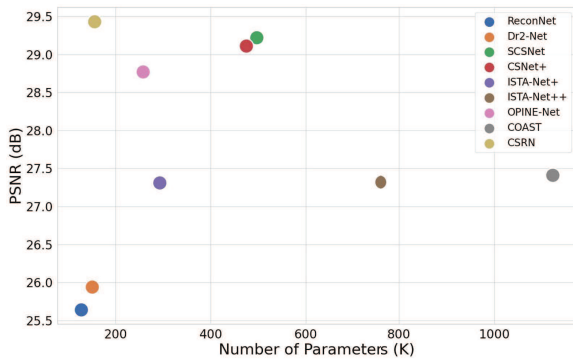}
    \caption{Comparison of the PNSRs and parameters of different deep learning-based methods on Set14 for a sample ratio of 0.1.}
    \label{psnr_parameter}
\end{figure}

\textbf{Parameter comparison: } Furthermore, we count each model's parameters and list the results in Table \ref{parameter}. For models that include sampling and reconstruction components, like SCSNet, CSNet+, ISTA-Net\textsuperscript{+}, ISTA-Net\textsuperscript{++},  COAST, OPINE-Net and CSRN, we count the reconstruction component's parameters. As shown, DPA-Net has the most parameters under all sample ratios, followed by COAST, while ReconNet has the fewest parameters when the sample ratio is less than 0.1; CSRN has the fewest parameters when the sample ratio is higher than 0.1. In fact, precisely because the progressive initial reconstruction sub-network constantly reconstructs the image by fusing feature maps, at high sample ratios, our initial reconstruction module can significantly reduce the number of parameters. Compared with other methods, CSRN uses progressive initial reconstruction and a recurrent structure, so it does not need as many parameters. While CSRN has more parameters than ReconNet and Dr\textsuperscript{2}-Net for sample ratios less than 0.1, it has a better reconstruction quality; Fig. \ref{psnr_parameter} provides an intuitive illustration of this. As a result, our proposed CSRN achieves a good balance between the number of parameters and the quality of the reconstructions. In other words, CSRN not only achieves a better reconstruction quality but also has a competitive reconstruction speed and a relatively small number of parameters.

\subsection{Analysis of Progressive Initial Reconstruction}

\begin{table}[t]
    \centering
    \setlength{\belowcaptionskip}{0.3cm}
    \caption{\centering Time Consumption Comparison of Different Methods}
    \renewcommand\arraystretch{1}
    \begin{tabular}{c|c|c}
        \hline
        \hline
        Algorithm & Time (s) & Device\\
        \hline
        DWT & 6.2973 & Intel Core i7-7770\\
        \hline
        MH & 6.6910 & Intel Core i7-7770\\
        \hline
        GSR & 821.2338 & Intel Core i7-7770\\
        \hline
        ReconNet & \textbf{0.0013} & NVIDIA Titan Xp\\
        \hline
        Dr\textsuperscript{2}-Net & \textcolor{blue}{0.0016} & NVIDIA Titan Xp\\
        \hline
        DPA-Net &  0.0311 & NVIDIA Titan Xp\\
        \hline
        SCSNet & 0.0283 & NVIDIA Titan Xp\\
        \hline
        CSNet\textsuperscript{+} &  0.0025 & NVIDIA Titan Xp\\
        \hline
        ISTA-Net\textsuperscript{+} &  0.0071 & NVIDIA Titan Xp\\
        \hline
        ISTA-Net\textsuperscript{++} &  0.0123 & NVIDIA Titan Xp\\
         \hline
         COAST & 0.0163 & NVIDIA Titan Xp\\
        \hline
        OPINE-Net & 0.0209 & NVIDIA Titan Xp\\
        \hline
        CSRN &  0.0022 & NVIDIA Titan Xp\\
        \hline
        \hline
    \end{tabular}
    \label{time_comparison}
\end{table}

\begin{table}[t]
    \centering
    \setlength{\belowcaptionskip}{0.3cm}
    \caption{\centering Comparison of the Number of Parameters (K) of Deep Learning CS Networks}
    \renewcommand\arraystretch{1}
    \begin{tabular}{c|c|c|c|c|c|c|c}
        \hline
        \hline
		\multirow{2}{*}{Algorithm} & \multicolumn{7}{c}{Sample Ratio}\\
		\cline{2-8}& 0.01 & 0.05 & 0.1 & 0.2 & 0.3 & 0.4 & 0.5\\
        \hline
        ReconNet & \textbf{34} & \textbf{76} & \textbf{128} & \textcolor{blue}{234} & \textcolor{blue}{338} & 444 & 548\\
        \hline
        Dr\textsuperscript{2}-Net & \textcolor{blue}{57} & \textcolor{blue}{99} & \textcolor{blue}{151} & 256 & 361 & 466 & 571\\
        \hline
        DPA-Net & 9310 & 9310 & 9310 & 9310 & 9310 & 9310 & 9310\\
        \hline
        SCSNet & 497 & 497 & 497 & 497 & 497 & 497 & 497\\
        \hline
        CSNet\textsuperscript{+} & 381 & 423 & 475 & 579 & 685 & 789 & 895\\
        \hline
        ISTA-Net\textsuperscript{+} & 176 & 259 & 293 & 433 & 572 & 712 & 852\\
         \hline
        ISTA-Net\textsuperscript{++} & 760 & 760 & 760 & 760 & 760 & 760 & 760\\
        \hline
        COAST & 1122 & 1122 & 1122 & 1122 & 1122 & 1122 & 1122\\
        \hline
        OPINE-Net & 205 & 231 & 258 & 287 & 349 & \textcolor{blue}{402} & \textcolor{blue}{414}\\
        \hline
        CSRN & 132 & 143 & 156 & \textbf{193} & \textbf{231} & \textbf{268} & \textbf{306}\\
        \hline
        \hline
    \end{tabular}
    \label{parameter}
\end{table}

In our experiments, the simple initial reconstruction process uses a simple convolutional layer to perform dimensional mapping. Our initial reconstruction sub-network has a hierarchical structure and progressively recovers the original images, meaning that it needs fewer parameters for initial reconstructions. Table \ref{parameter_comparison} compares the parameters needed for simple initial reconstruction and progressive initial reconstruction. When the sample ratio is 0.01, progressive initial reconstruction requires slightly more parameters than simple initial reconstruction. At high sample ratios, however, progressive initial reconstruction requires significantly fewer parameters. As an example, progressive initial reconstruction requires one third of the number of parameters required by simple initial reconstruction when the sample ratio is 0.5.

To investigate the effect of progressive initial reconstruction on the quality of the initial reconstructed image, we replace it with simple initial reconstruction and test the quality of the initial reconstructed images from DIV2K; the results of this comparison are listed in Table \ref{initial_quality_comparison}. It can be observed that progressive initial reconstruction can achieve a better quality at low sample ratios, but its reconstruction quality is not more advantageous at high sample ratios. This is partly due to the fact that stacking too many basic reconstruction blocks at high sample ratios makes gradient propagation difficult during training. In general, the results demonstrate that progressive initial reconstruction achieves a competitive reconstruction quality when it comes to reconstructing initial images with a smaller number of parameters.

\subsection{Analysis of RRFM}

\begin{figure*}[t]
    \centering
    \includegraphics[width=0.95\textwidth]{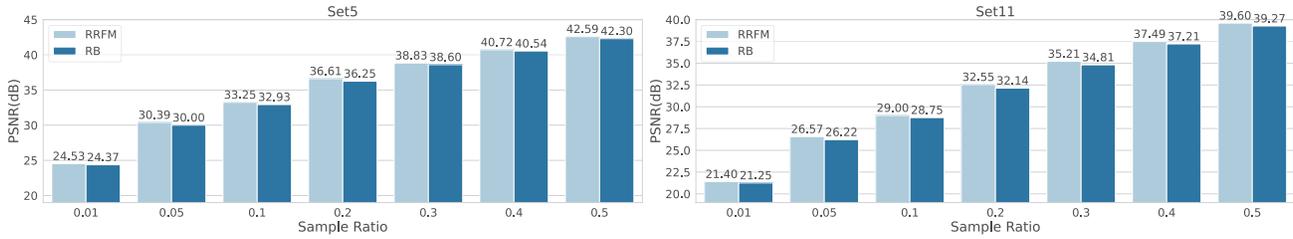}
    \caption{Reconstruction quality comparison of the RRFM and RB.}
    \label{rrfm_test}
\end{figure*}

The RRFM is characterized by multi-scale feature learning and a recurrent structure; it can recurrently process input features through a residual block and fuse the features of different receptive fields. To investigate the effect of the two parts of the RRFM, we remove the recurrent connection from the RRFM and the convolutional layer at the end of the RRFM, and thus, the RRFM is reduced to a normal residual block (RB). We compare the reconstruction qualities of CSRN with the RRFM and RB on Set5 and Set11, and the results are given in Fig. \ref{rrfm_test}. As shown, without the feature fusion operation and recurrent connection, the reconstruction quality of CSRN obviously decreases. This suggests that the feature fusion operation conducted using 1*1 convolution is helpful for the comprehensive utilization of the multi-scale feature maps generated during recurrences. Additionally, this means that the recurrent connection can enhance the network's ability to explore the useful information in feature maps.

Thanks to the recurrent structure of the RRFM, the number of parameters is reduced and the cost of training the neural network model is thus alleviated. It is also worth noting that although the performance of CSRN with the RB is not as good as that of CSRN with the RRFM, it is still comparable to CSNet\textsuperscript{+}. This demonstrates the effectiveness of the overall structural design of CSRN. Due to the absence of the recurrent structure, the computational requirements of CSRN with the RB decrease, and thus it can recover images with a smaller memory footprint compared to normal CSRN. When limited computing resources are available, we can use CSRN with the RB to recover images with a relatively good quality.

\begin{table}[t]
    \setlength{\belowcaptionskip}{0.3cm}
    \caption{Comparison of the Parameters Needed for Simple Initial Reconstruction and Progressive Initial Reconstruction}
    \centering
    \renewcommand\arraystretch{1}
    \begin{tabular}{c|c|c}
        \hline
        \hline
        \multirow{2}{*}{Sample ratio}
        & Simple initial & Progressive initial\\
        & reconstruction (K)& reconstruction (K)\\
        \hline
        0.01 & \textbf{10.2} & 13.9\\
        \hline
        0.05 & 52.2 & \textbf{24.4}\\
        \hline
        0.1 &  104.4 & \textbf{37.5}\\
        \hline
        0.2 &  208.9 & \textbf{75}\\
        \hline
        0.3 &  313.3 & \textbf{112.5}\\
        \hline
        0.4 &  417.8 & \textbf{150}\\
        \hline
        0.5 &  522.2 & \textbf{187.6}\\
        \hline
        \hline
    \end{tabular}
    \label{parameter_comparison}
\end{table}

\begin{table}[t]
    \setlength{\belowcaptionskip}{0.3cm}
    \caption{Reconstruction Quality Comparison of Simple Initial Reconstruction and Progressive Initial Reconstruction}
    \centering
    \renewcommand\arraystretch{1}
    \begin{tabular}{c|c|c}
        \hline
        \hline
        \multirow{2}{*}{Sample ratio}
        & Simple initial & Progressive initial\\
        & reconstruction (dB)& reconstruction (dB)\\
        \hline
        0.01 & 24.53 & \textbf{24.74}\\
        \hline
        0.05 & 28.14 & \textbf{28.31}\\
        \hline
        0.1 &  30.19 & \textbf{30.28}\\
        \hline
        0.2 &  32.92 & \textbf{33.01}\\
        \hline
        0.3 &  35.19 & \textbf{35.22}\\
        \hline
        0.4 &  37.21 & \textbf{37.23}\\
        \hline
        0.5 &  \textbf{39.33} & 39.22\\
        \hline
        \hline
    \end{tabular}
    \label{initial_quality_comparison}
\end{table}

\subsection {Analysis of Feature Compression Strategy}

\begin{figure*}[t]
    \centering
    \includegraphics[width=0.95\textwidth]{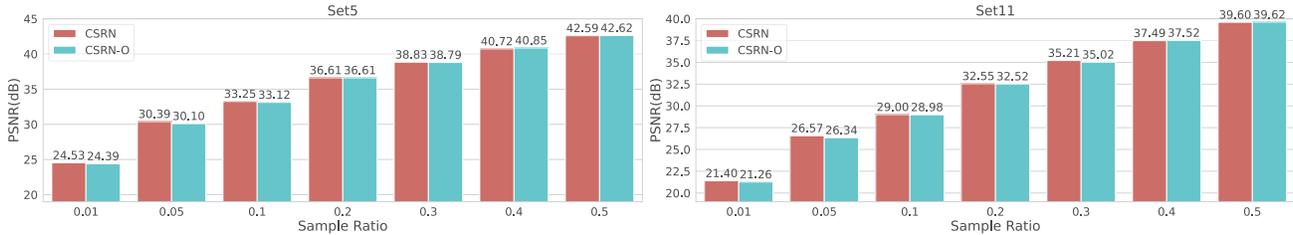}
    \caption{Reconstruction quality comparison of CSRN and CSRN-O.}
    \label{csrn_compress}
\end{figure*}

\begin{figure*}[t]
    \centering
    \includegraphics[width=0.95\textwidth]{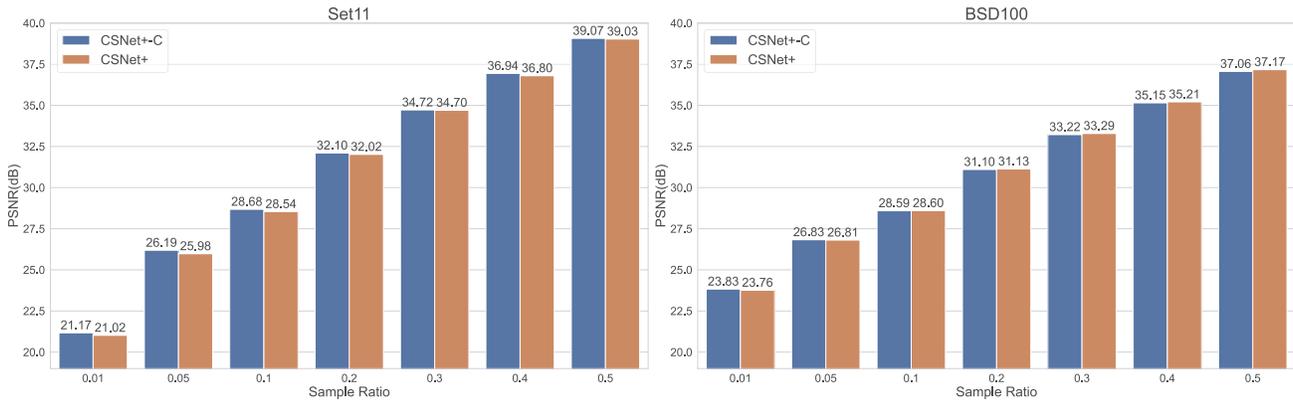}
    \caption{Reconstruction quality comparison of CSNet\textsuperscript{+} and CSNet\textsuperscript{+}-C.}
    \label{csnet_compress}
\end{figure*}

In this subsection, we investigate the effect of the feature compression strategy on the network performance in terms of the reconstruction quality, generality and resource consumption of the model. 

To test the effect of the compression strategy on the reconstruction quality, we replace $ Conv(m, 2, 2) $ at the head of the residual reconstruction sub-network with $ Conv(m, 3, 1) $, and thus the feature maps will retain the resolution of the original image during the inference process. We denote this model as CSRN-O and compare its performance with CSRN on the Set5 and Set11. As shown in Fig. \ref{csrn_compress}, CSRN can achieve a performance comparable to that of CSRN-O. As the sample ratio further increases, CSRN-O performs better. This can be explained by the fact that in the case of a low sample ratio, some information contained in the feature maps $F_i$ from the initial reconstruction sub-network is redundant for residual image reconstruction, and thus compressing the feature maps helps the network to extract meaningful information. As the sample ratio further increases, more information can be preserved during the sampling process, and $ F_i$ can contain more useful information for residual images. In this case, the compression operation may cause some information loss and CSRN’s performance may decrease slightly. As we only use a relatively simple strategy for compression, a better-designed method may be able to eliminate such information loss. 

To verify the generality of the compression strategy, we apply it to CSNet\textsuperscript{+} by replacing $ Conv(64, 3, 1) $ at the head of the deep reconstruction part with $ Conv(64, 2, 2) $, and we replace $ Conv(64, 3, 1) $ at the tail of the deep reconstruction part with $ PixelShuffle(2) $ and $ Conv(16, 3, 1) $. We denote CSNet\textsuperscript{+} with the compression strategy as CSNet\textsuperscript{+}-C and compare it with the original CSNet\textsuperscript{+}. The results are shown in Fig. \ref{csnet_compress}; it can be seen that like CSRN, CSNet\textsuperscript{+}-C can achieve a better reconstruction quality at low sample ratios. As the sample ratio increases, the reconstruction quality of CSNet\textsuperscript{+} approaches that of CSNet\textsuperscript{+}-C and finally surpasses that of CSNet\textsuperscript{+}-C. Therefore, CSNet\textsuperscript{+}-C can achieve a reconstruction quality that is comparable to that of CSNet\textsuperscript{+}, which proves that the feature compression strategy has a certain degree of generalizability.

\begin{figure}[t]
    \centering
    \includegraphics[width=3.0in]{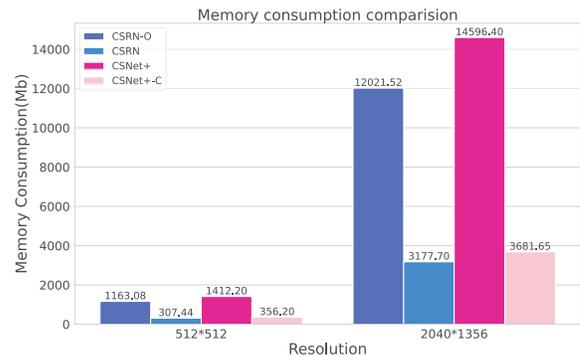}
    \caption{Comparison of the effects of the compression strategy on memory consumption.}
    \label{memory_consumption}
\end{figure}

We test the memory consumption during the feedforward process and the reconstruction speed to explore the effect of the compression strategy on the model's resource consumption. We send images with resolutions of $512 \times 512$ and $2040 \times 1356$ into the abovementioned models to test their memory consumption, and the results are shown in Fig. \ref{memory_consumption}. As expected, the compression strategy reduces the size of the feature maps, and thus it can significantly reduce the memory consumption, especially when high-resolution images are utilized.

\subsection{Summary of the Three Proposed Strategies}
We have independently analyzed the three strategies proposed in this paper, which include the RRFM, progressive initial reconstruction and feature compression strategy. In this subsection, we will summarize the advantages of each strategy. The feature compression strategy reduces the size of the feature map to one fourth of the size of the original map, and this reduces the number of parameters and decreases the memory consumption even more than progressive initial reconstruction. The RRFM is the most important part of our model; it significantly reduces reconstruction errors by using residual reconstruction to preserve reconstruction details.

\section{Conclusion}
In this paper, we propose a sustainable image compressed sensing method that can be directly used for image sampling and recovery. The method is both lightweight and efficient, and it can achieve a better performance with less memory consumption, a smaller number of parameters and competitive reconstruction time. Recently, neural networks with physics-driven structures for image compressed sensing have attracted more attention. These networks unfold optimization steps using neural networks, which enables some prior knowledge to be implicitly contained in the model. As a result, these networks have achieved a certain degree of interpretability and thus have the potential to be used in our future research.


\ifCLASSOPTIONcaptionsoff
  \newpage
\fi



%

\bibliographystyle{IEEEtran}
\bibliography{ref_clean}

%

\begin{IEEEbiography}[{\includegraphics[width=1in,height=1.25in,clip,keepaspectratio]{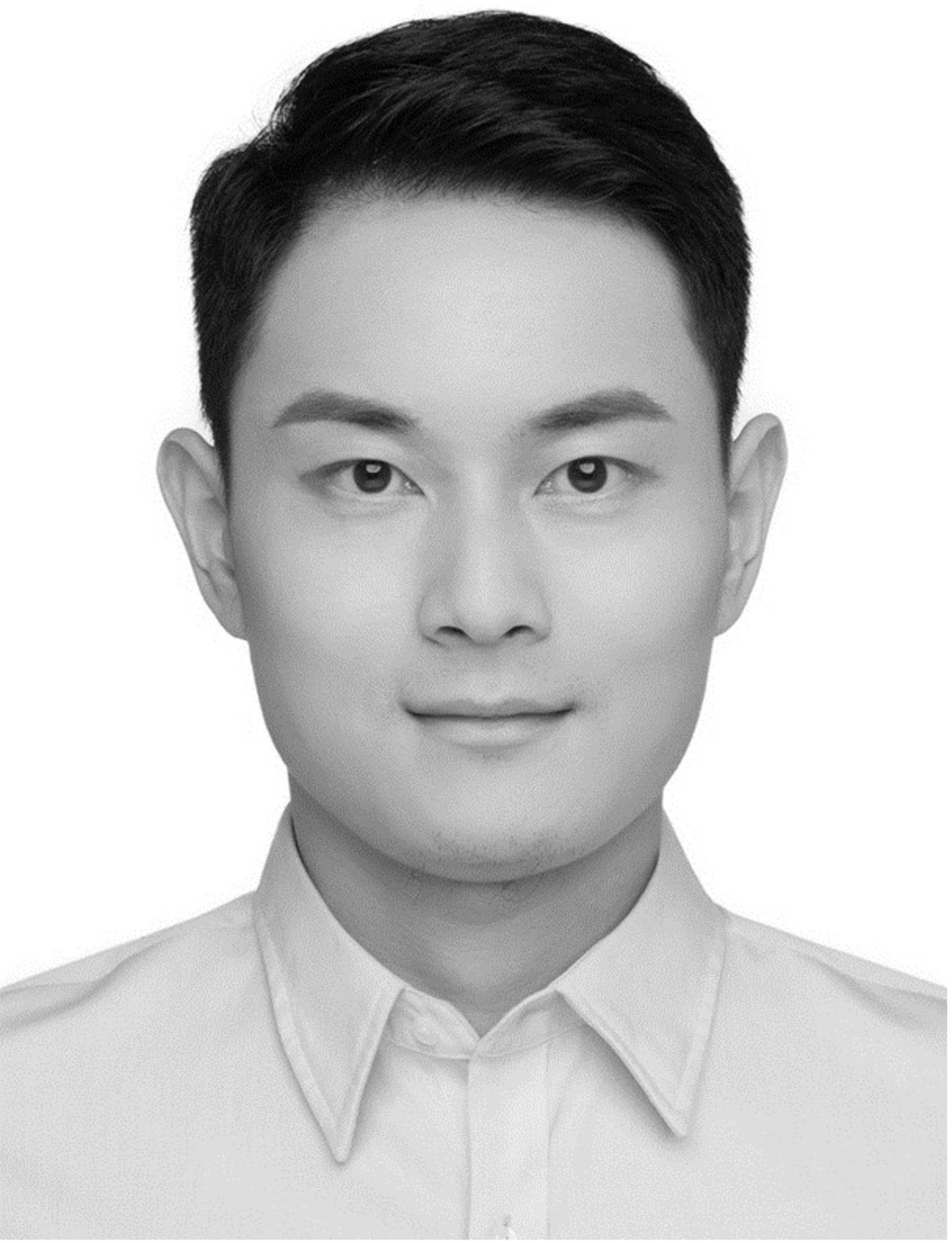}}]{Yu Zhou}
received the B.Sc. degree in electronics and information engineering and the M.Sc. degree in circuits and systems from Xidian University, Xi’an, China, in 2009 and 2012, respectively, and the Ph.D. degree in computer science from the City University of Hong Kong, Hong Kong, in 2017. He is currently an Assistant Professor with College of Computer Science and Software Engineering, Shenzhen University, Shenzhen, China. His current research interests include computational intelligence, machine learning and intelligent information processing. He is an IEEE member.
\end{IEEEbiography}
\begin{IEEEbiography}[{\includegraphics[width=1in,height=1.25in,clip,keepaspectratio]{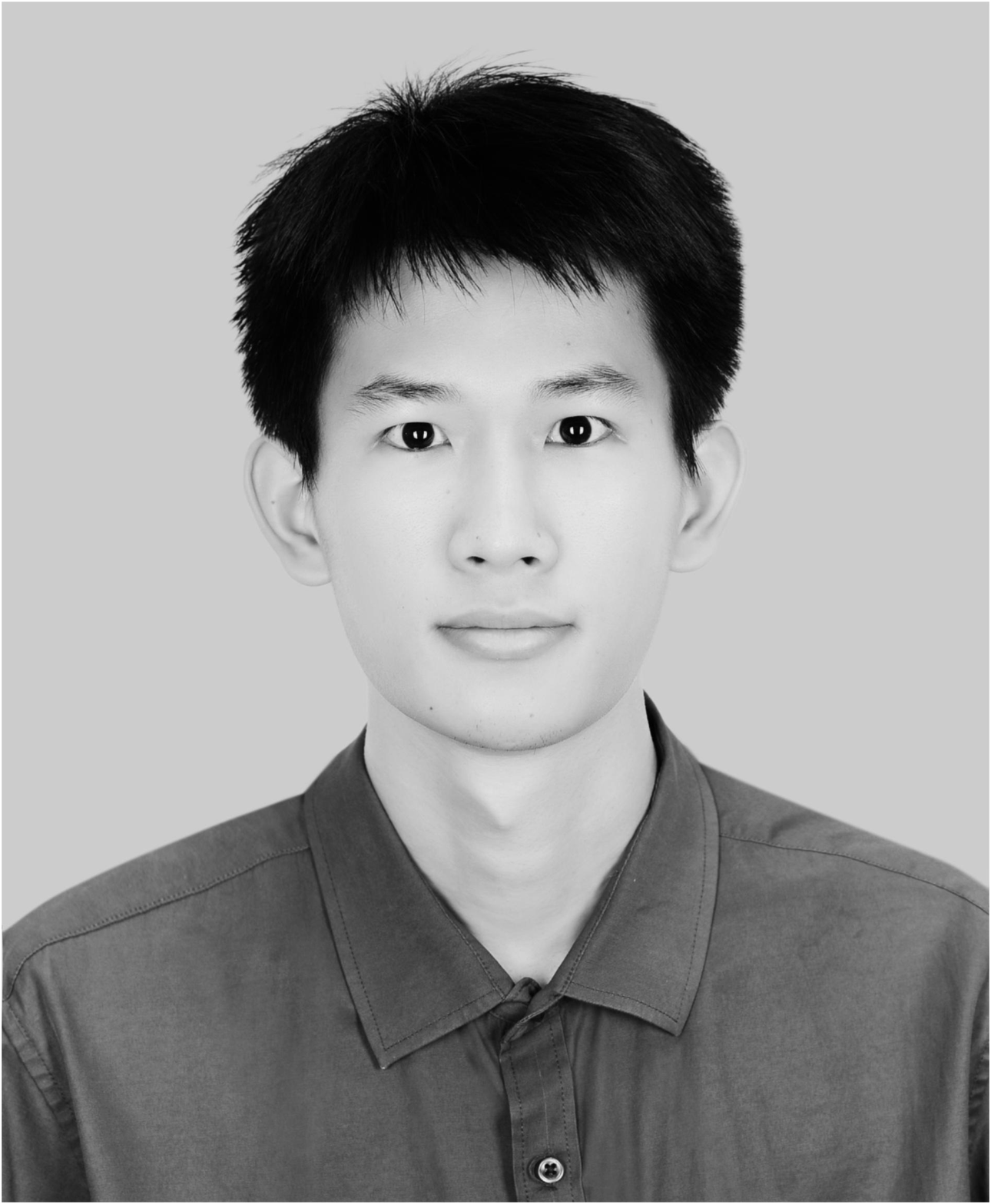}}]{Yu Chen}
received the B.Eng. degree from East China University of Technology, JiangXi, China, in 2020. He is currently pursuing the M.Sc. degree at the College
of Computer Science and Software Engineering,
Shenzhen University, Shenzhen, China. His current research areas are compressed sensing and machine learning.
\end{IEEEbiography}
\begin{IEEEbiography}[{\includegraphics[width=1in,height=1.25in,clip,keepaspectratio]{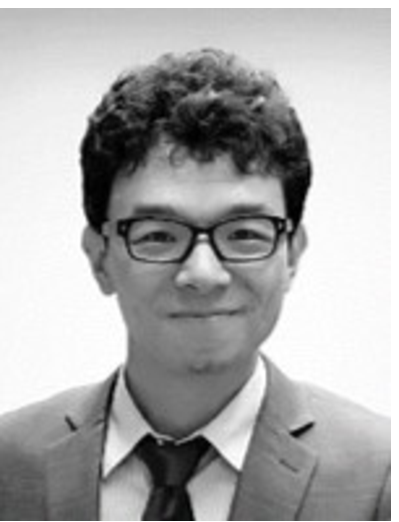}}]{Xiao Zhang}
received the B.Eng. and M.Eng. degrees from South-Central University for Nationalities, Wuhan, China, in 2009 and 2011 respectively. In 2015, he was a visiting scholar with the Utah State University, Utah, USA. He received his Ph.D. degree from Department of Computer Science in City University of Hong Kong, Hong Kong, 2016. During 2016-2019, he was a postdoc research fellow at Singapore University of Technology and Design. Currently, he is Associate Professor with College of Computer Science, South-Central University for Nationalities, China. His research interests include algorithms design and analysis, combinatorial optimization, wireless and UAV networking.
\end{IEEEbiography}

\begin{IEEEbiography}[{\includegraphics[width=1in,height=1.25in,clip,keepaspectratio]{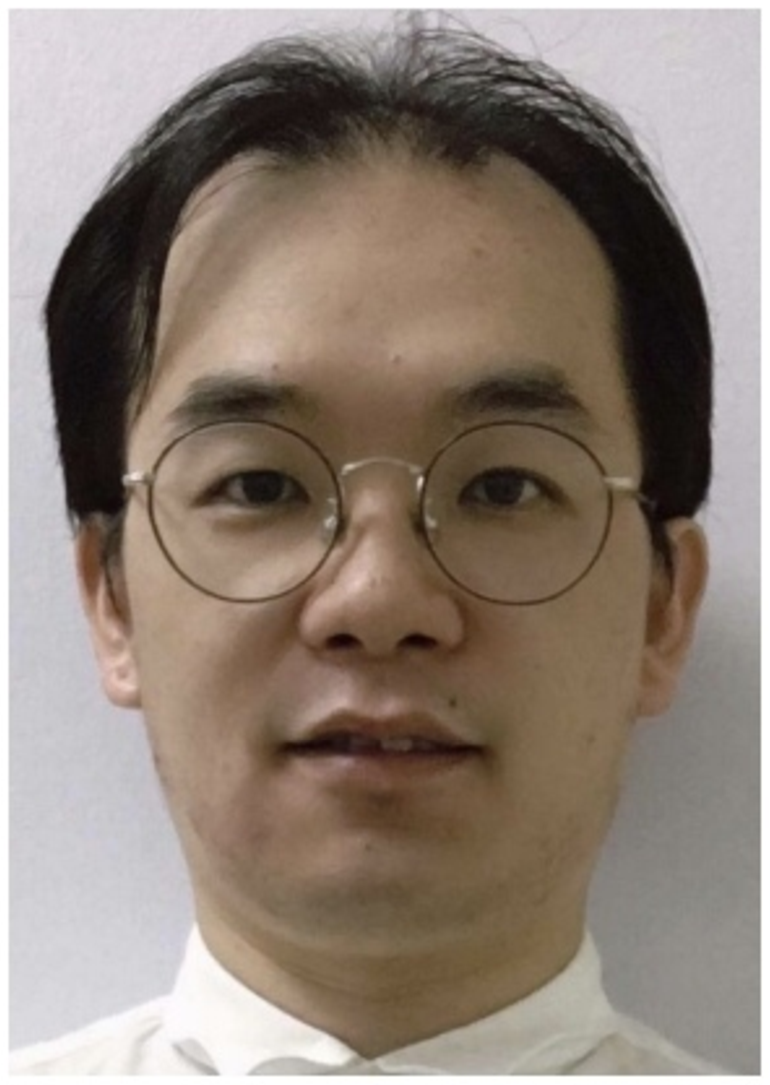}}]{Pan Lai}
    received the Ph.D. degree
from the School of Computer Engineering, Nanyang
Technological University, in 2016. He is currently an Associate Professor with South-Central University for Nationalities, China. He was a Post-Doctoral Research Fellow with the Singapore
University of Technology and Design, Singapore,
from 2016 to 2019. His research interests include
resource allocation and scheduling algorithm design in computer and network systems, network economics, and game theory.
\end{IEEEbiography}

\begin{IEEEbiography}[{\includegraphics[width=1in,height=1.25in,clip,keepaspectratio]{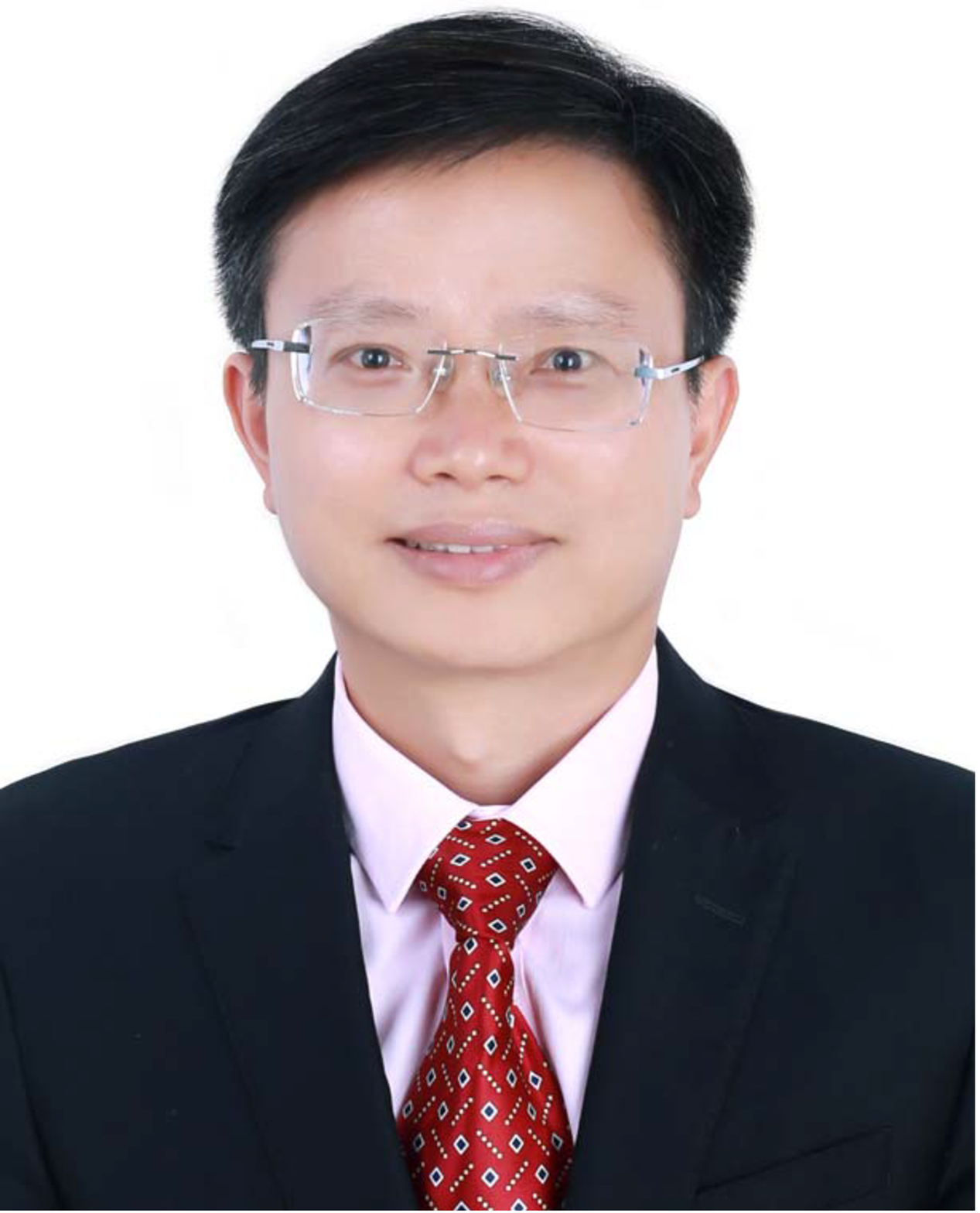}}]{Lei Huang}
    (Senior Member, IEEE)    received the
    B.Sc. and Ph.D. degrees in electronic engineering
    from Xidian University, Xi’an, China, in 2000 and
    2005, respectively. Since 2014, he has been with
    the College of Electronics and Information Engineering, Shenzhen University, where he is currently
    a Distinguished Professor and the Director of the
    Shenzhen Key Laboratory of Advanced Navigation
    Technology. He has undertaken eight projects from
    the National Natural Science Foundation of China
    (NSFC), including the National Science Funds for
    Distinguished Young Scholar. His research interests include array signal
    processing, statistical signal processing, sparse signal processing, and their
    applications in radar, navigation, and wireless communications. In 2018,
    he was elected as a fellow of IET. Since 2016, he has been an Elected Member
    of the Sensor Array and Multichannel (SAM) Technical Committee of the
    IEEE Signal Processing Society. He has been serving as an Associate Editor
    for the IEEE TRANSACTIONS ON SIGNAL PROCESSING since 2015. He is
    currently a Senior Area Editor of the IEEE TRANSACTIONS ON SIGNAL
    PROCESSING.
\end{IEEEbiography}

\begin{IEEEbiography}[{\includegraphics[width=1in,height=1.25in,clip,keepaspectratio]{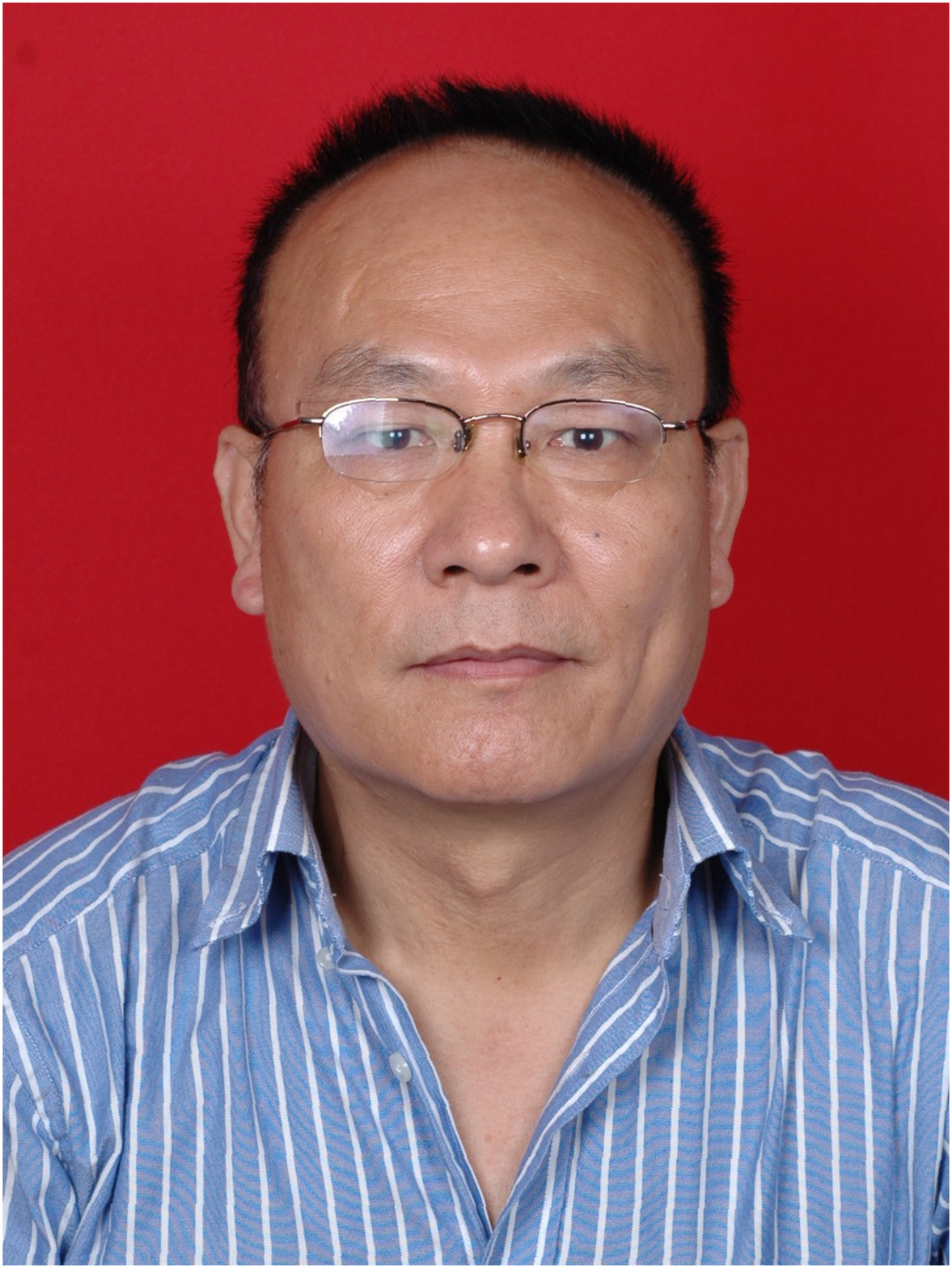}}]{jianmin Jiang}
    received PhD from the University of Nottingham, UK, in 1994. From 1997 to 2001, he worked as a full professor of Computing at the University of Glamorgan, Wales, UK. In 2002, he joined the University of Bradford, UK, as a Chair Professor of Digital Media, and Director of Digital Media Systems Research Institute. He worked at the University of Surrey, UK, as a full professor during 2010-2015 and a distinguished chair professor (1000-plan) at Tianjin University, China, during 2010-2013. He is currently a Distinguished Chair Professor and director of the Research Institute for Future Media Computing at the College of Computer Science Software Engineering, Shenzhen University, China. He was a chartered engineer, fellow of IEE(IET), fellow of RSA, member of EPSRC College in the UK, and EU FP-6/7 evaluator. His research interests is primarily focused on deep learning applications in multimedia content understanding and analysis, brain perceived pattern recognition and intelligent image/video processing. He has published around 400 refereed research papers in international leading journals and conferences.
\end{IEEEbiography}








\end{document}